\pdfoutput=1

\documentclass[10pt, journal,twoside]{IEEEtran}

\ifCLASSINFOpdf
\else
\fi

\usepackage{cite}
\usepackage{amsmath}
\usepackage{amsfonts}
\usepackage{algorithm}
\usepackage{algorithmic}
\usepackage{threeparttable}
\usepackage{graphicx}
\usepackage{subfigure}
\usepackage{float}
\usepackage{booktabs}
\usepackage{makecell}
\usepackage{multirow}
\usepackage{pifont}
\usepackage{amssymb}
\usepackage{mathrsfs}
\usepackage{tablefootnote}
\usepackage{soul}
\usepackage{ulem}

\usepackage{comment}
\usepackage{url}
\usepackage[table,xcdraw,dvipsnames]{xcolor}  
\usepackage{color}
\definecolor{citecolor}{RGB}{66,168,235}
\definecolor{linkcolor}{RGB}{255,0,0}
\usepackage{hyperref}
\hypersetup{colorlinks=true,citecolor=citecolor,linkcolor=linkcolor,urlcolor=Thistle}

\usepackage{colortbl}

\begin{document}

%
\title{
Advancing Plain Vision Transformer Towards Remote Sensing Foundation Model
}
%
%
%

\author{Di Wang,
        Qiming Zhang,
        Yufei Xu,
        Jing Zhang,~\IEEEmembership{Member,~IEEE,}
        Bo Du,~\IEEEmembership{Senior Member,~IEEE,}\\
        Dacheng Tao,~\IEEEmembership{Fellow,~IEEE}
        and Liangpei Zhang,~\IEEEmembership{Fellow,~IEEE}

\thanks{D. Wang, Q. Zhang and Y. Xu contributed equally to this paper. \textit{(Corresponding author: Bo Du)}}
\thanks{D. Wang and B. Du are with the School of Computer Science, Wuhan University, Wuhan 430072, China (e-mail: wd74108520@gmail.com; dubo@whu.edu.cn). B. Du is also with the National Engineering Research Center for Multimedia Software and Institute of Artificial Intelligence, Wuhan University, Wuhan 430072, China.}
\thanks{Q. Zhang, Y. Xu and J. Zhang are with the School of Computer Science, Faculty of Engineering, The University of Sydney, Australia (e-mail: qzha2506@uni.sydney.edu.au; yuxu7116@uni.sydney.edu.au; jing.zhang1@sydney.edu.au).}
\thanks{D. Tao is with the JD Explore Academy, China and is also with the School of Computer Science, Faculty of Engineering, The University of Sydney, Australia (e-mail: dacheng.tao@gmail.com).}
\thanks{L. Zhang is with the State Key Laboratory of Information Engineering in Surveying, Mapping and Remote Sensing, Wuhan University, Wuhan 430079, China (e-mail: zlp62@whu.edu.cn).}
}

%
%

\markboth{Journal of \LaTeX\ Class Files,~Vol.~14, No.~8, August~2015}{Wang \MakeLowercase{\textit{et al.}}: ADVANCING PLAIN VISION TRASFORMER TOWARDS RS FOUNDATION MODEL}
%



\maketitle

\begin{abstract}

  Large-scale vision foundation models have made significant progress in visual tasks on natural images, with vision transformers being the primary choice due to their good scalability and representation ability. However, large-scale models in remote sensing (RS) have not yet been sufficiently explored. In this paper, we resort to plain vision transformers with about 100 million parameters and make the first attempt to propose large vision models tailored to RS tasks and investigate how such large models perform. To handle the large sizes and objects of arbitrary orientations in RS images, we propose a new rotated varied-size window attention to replace the original full attention in transformers, which can significantly reduce the computational cost and memory footprint while learning better object representation by extracting rich context from the generated diverse windows. Experiments on detection tasks show the superiority of our model over all state-of-the-art models, achieving 81.24\% mAP on the DOTA-V1.0 dataset. The results of our models on downstream classification and segmentation tasks also show competitive performance compared to existing advanced methods. Further experiments show the advantages of our models in terms of computational complexity and data efficiency in transferring. The code and models will be released at \href{https://github.com/ViTAE-Transformer/Remote-Sensing-RVSA}{https://github.com/ViTAE-Transformer/Remote-Sensing-RVSA}

\end{abstract}


%
\IEEEpeerreviewmaketitle

\section{Introduction}
%
%
%
%

\IEEEPARstart{I}N the geomatics community, the remote sensing image (RSI) is an important data source for earth observation since it is easy to access, can be obtained in real-time, and provides abundant geospatial and spectral information. The RSIs are being employed in many valuable applications such as scene recognition for land use and land cover classification for precision agriculture \cite{agriculture_1,zhang2020empowering} and object detection for maritime monitoring \cite{app_det1}. 

It is necessary to effectively represent the RSI contents and attributes to perform the above applications. \cite{zyx_1,asr_2021_jstars_f2brbm,zyx_2,ass_2022_tgrs_factseg,NN_TSP_object_detection,zyx_3,acd_2021_jstars_dasnet} In the current remote sensing (RS) field, convolutional neural networks (CNNs) are the most commonly used models for extracting hierarchical multiscale visual features. \cite{xu2021_dfagcn,farseg,acd_2021_grsl_snunet,aod_2022_tgrs_s2anet} However, existing studies \cite{erf,segformer} reveal that the limited receptive field of convolution in each layer makes it difficult for CNNs to pay attention to long-range pixels and extract global context. To address this issue, the self-attention (SA) mechanism \cite{non-local} is proposed to obtain flexible global dependency by enabling the interaction between arbitrary pixels in images, delivering promising results in the computer vision (CV) field. Further, vision transformer \cite{vit} adopts the design of multi-head SA (MHSA), which simultaneously implements the above procedure in multiple projected subspaces, which diversifies the extracted contexts and improves the feature representation.

\begin{figure}[t]
  \centering
  \includegraphics[width=1\linewidth]{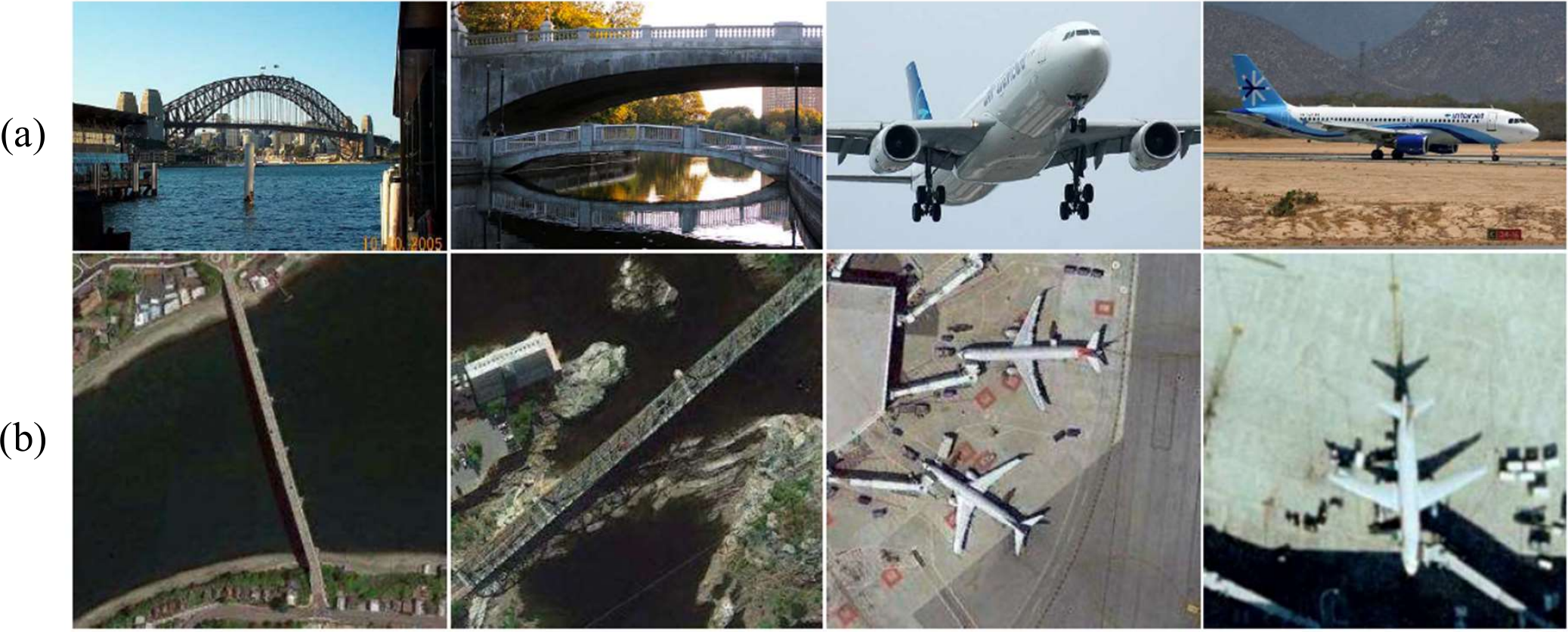}\\
  \caption{The comparison between (a) natural images and (b) RSIs. Here, we show some common categories, including bridge and airplane, in both types of images. Compared with the natural images from horizontal observation, RSIs tend to be in an overhead view. The natural images are from the ImageNet-1K dataset, while the RSIs are chosen from the UCM, AID, or NWPU datasets.
  }
  \label{sample_images}
\end{figure}

The plain vision transformer (ViT) \cite{vit} is a straightforward architecture that stacks several transformer encoder blocks sequentially after the patch embedding layer, where features after each block are at the same scale. To better adapt the vision transformers to downstream tasks, researchers borrow the idea of hierarchical design in CNNs and devise hierarchical vision transformers accordingly \cite{swint,vitae_v2}. These models are usually pretrained in a supervised way using large-scale datasets and finetuned on the downstream tasks. The recent work \cite{wang_rsp_2022} empirically studies the hierarchical vision transformers for RSIs by comparing different pretraining strategies and models. It confirms the superiority of hierarchical vision transformers over CNNs and reveals the effectiveness of pretraining with RS scene datasets like MillionAID \cite{Long2021DiRS}. Nevertheless, is the hierarchical structure a must for better performance in RS tasks? To find the answer, this paper makes the first attempt in this direction by employing the non-hierarchical plain structure. Thanks to the development of the unsupervised learning in masked image modeling (MIM) \cite{mae}, recent work reveals that such a pretraining process can give a good initialization for plain ViTs to achieve surprising results on various downstream tasks including object detection and pose estimation \cite{vitdet,vitpose}. For example, ViTDet \cite{vitdet} uses intermediate features at the same scale and upsamples/downsamples them to build the feature pyramid for object detection. The main insight behind their success is that the multiscale prior can be learned from the data during the pretraining, thus making it possible to discard the hierarchical structure.

Inspired by the above pilot studies, we employ the same MIM pretraining and finetuning routine to investigate the influences of plain ViTs on RS tasks. Usually, obtaining the annotations is expensive and nontrivial since understanding RSIs usually requires some expert experience. By contrast, it is much easier to obtain massive unlabeled RSIs in different resolutions and temporals from different kinds of sensors, e.g., via numerous satellites for continuous earth observations. How to leverage the abundant unlabeled RSIs for pretraining has become an active research topic in the RS community. A few self-supervised learning (SSL) methods \cite{seco, geography_aware} have been proposed, which, however, only target CNNs and have not been proven effective for large-scale vision transformer models. Recently, reconstruction-based SSL methods such as MAE \cite{mae} have been proposed and shown effective for pretraining plain ViTs and adapting them for downstream tasks~\cite{vitdet,vitpose}. In this paper, we also employ MAE to pretrain the plain ViT and the recently proposed ViTAE transformer with about 100M parameters on the MillionAID dataset without using the labels.

After pretraining, we adapt the vision transformers to downstream tasks via finetuning on the corresponding datasets. To reduce the computational cost and memory footprint, a natural choice is to replace the full self-attention with local window attention \cite{vitdet}. However, the windows are in fixed sizes and locations, which may restrict the region to extract useful context and limit the model representation ability. A recently proposed method named varied-size attention (VSA) \cite{zhang2022vsa} addresses this issue by learning trainable scaling factors and offsets to adapt the windows' size, shape, and location to diverse image content, thus delivering better performance on many tasks. However, the objects in natural images are generally oriented upward due to gravity, while those in RSIs usually appear with various orientations as shown in Figure \ref{sample_images}. To handle this difference, we propose to extend VSA to rotated varied-size attention (RVSA). It introduces an extra learnable rotation mechanism in determining the window configurations where oriented windows at different angles, sizes, shapes, and locations are obtained to extract richer context. We evaluate the proposed method on both plain ViT \cite{vit} and ViTAE \cite{vitae_v2} models for three kinds of RS tasks including scene classification, semantic segmentation, and object detection. We hope this study can fill the gap and provide useful insights about developing large plain ViTs for the RS community.

The main contribution of this study is three-fold:
\begin{itemize}
  \item[(1)] We demonstrate the possibility of pretraining plain ViTs with about 100 million parameters on RSIs, adapting them for downstream RS tasks, and delivering competitive performance. To our best knowledge, they are so far the largest models in the RS community, making a step towards the RS foundation model \cite{yu2022coca, yuan2021florence, bommasani2021opportunities, ringmo} of impressive representation ability.
  \item[(2)] We introduce a learnable rotation mechanism into the vision transformer to learn varied-size windows with different orientation angles for attention calculation, which is very suitable for dealing with RSIs. It promotes the extraction of rich context from the generated windows and learning better feature representations.
  \item[(3)] Experimental results show that the proposed models set new state-of-the-art (SOTA) on the detection task, and obtain competitive performances on classification and segmentation tasks. Besides, we also show the advantages of the proposed models in terms of computational complexity and data efficiency in transferring.
 \end{itemize}

The remainder of this paper is organized as follows. Section II briefly reviews related works including vision foundation models, window-based vision transformers, and model pretraining for RS tasks. Section III describes the proposed method, where we separately present the implementation of unsupervised MAE pretraining on MillionAID and the proposed RVSA method. The experiment results on the three tasks and the related comprehensive analyses are presented in Section IV. Finally, Section V concludes this paper.

\section{Related Work}
\subsection{Vision Foundation Model}
Foundation models based on transformers have demonstrated strong capabilities in both vision and language tasks~\cite{bert,switch_transformer}. Vision transformers have also experienced rapid development towards large-scale foundation models thanks to their great potential in scalability and structure flexibility, e.g., the model size can be easily scaled up via stacking the same transformer layers with widening dimension~\cite{scale_vit}, introducing a mixture of experts~\cite{vmoe}, and introducing inductive bias~\cite{coatnet,vitae_v2,swint_v2}. Among them, plain ViT structures have attracted more attention for their simplicity and flexibility in input formats and superior performance in natural image classification tasks~\cite{yu2022coca}. However, how to adapt the plain ViTs with isotropic structures to downstream visual tasks remains challenging. Recently, ViTDet~\cite{vitdet} demonstrates that although there is no multi-stage structure in plain ViTs, they can generate multi-scale features via simple up- and down-sampling modules and use the window-attention mechanism to reduce computational cost and memory footprint significantly. Similarly, ViTPose~\cite{vitpose} shows the potential of the plain structure on pose estimation tasks with a simpler decoder. Inspired by their success, we argue it is also of great significance to explore the potential of the plain ViT structures in RS tasks, which has been largely ignored. In this work, we develop the first plain ViT backbone networks for RS tasks and scale them up to 100M parameters, which are the largest models in RS literatures. Equipped with the proposed RVSA method, we show superior performance on different RS tasks including classification, detection, and segmentation.

\subsection{Window-Based Vision Transformer}
Although the plain ViTs demonstrate superior performance with model scaling, the full attention operation employed in them brings quadratic complexities over the image size, limiting their applications in downstream visual tasks where high-resolution images are ubiquitous. To address this issue, window-based MHSA (WMHSA)~\cite{swint} has been employed by partitioning the images into non-overlapping windows and conducting MHSA inside each window. WMHSA has linear computational complexity with respect to image size, making it possible to process high-resolution images. For example, ViTDet~\cite{vitdet} explores interleaved windows-based and full attention modules to process images with sizes up to $1,024 \times 1,024$. Despite the success of WMHSA in balancing memory consumption and performance, it needs extra mechanisms to bridge the connection between different windows, e.g., shifting operations or full attentions. To alleviate such an issue, some works introduce different window partitions with the connection between adjacent windows or additional tokens for cross-window information exchange \cite{cswin,pale}. Despite their efficiency in different vision tasks, they rely on hand-crafted designs of the window shape and size as well as sequentially stack these layers for modeling relationships between distant tokens. Recently, VSA~\cite{zhang2022vsa} proposes to learn adaptive window sizes in a data-driven manner, where the windows could cover different regions and promote cross-window information exchange. In this work, we extend the VSA idea to adapt the plain ViT for RS tasks. Specifically, to better adapt the vision transformer to RSIs, we propose the RVSA method by introducing a learnable rotation mechanism. Such a strategy provides network-oriented windows at different angles, sizes, shapes, and locations to learn better object representation. The orientation-aware mechanism improves the performance of vision transformers on different RS tasks. 
 
\subsection{Model Pretraining for RS task}
Model pretraining is essential in improving the performance of deep neural networks on RSIs. Previous methods mostly utilize the ImageNet~\cite{imagenet} dataset for pretraining and improve performance in classification, detection, and segmentation with task-specific designs~\cite{lse_2021_asr,roi_transformer,farseg}. Unfortunately, due to the huge difference between natural images and RSIs, there always exist domain gaps to be mitigated when transferring these models pretrained on natural images to RS tasks. Recently, \cite{wang_rsp_2022} proposes to pretrain deep models on a large-scale RS dataset named MillionAID in a fully-supervised manner. It demonstrates that the pretraining on RS datasets can help improve the performance of both CNNs and vision transformers. However, the requirement of more labeled data is still a burden for pretraining much larger models. To break the labeled data insufficiency restriction, RS representations from other resources are explored. For example, GeoKR \cite{geokr} leverages the land cover ratio in the digital geographical information product as a weak label for RS representation learning. However, the shift of label distribution between data from different resources still impedes learning. Some works resort to SSL~\cite{simclr, mocov2, regioncl, mae} with different RS characteristics taken into the design, e.g., exploring seasonal variants for constructing positive pairs~\cite{seco} or leveraging spatial and temporal information~\cite{geography_aware}. However, they only target CNNs. A very recent method RingMo \cite{ringmo} employs an advanced MIM method named SimMiM \cite{simmim} to pretrain large-scale vision transformer models. Different from these works, we use the representative MAE method~\cite{mae} for MIM pretraining and specifically focus on advancing plain ViTs for RS tasks.

\section{Proposed Method}
 
In this section, we will first introduce the details about the unsupervised pretraining of plain ViT and ViTAE with MAE and the design of the proposed RVSA modules. Then, we will briefly discuss how to transfer the pretrained models to different RS tasks.

 \begin{figure*}[t]
  \centering
  \includegraphics[width=\linewidth]{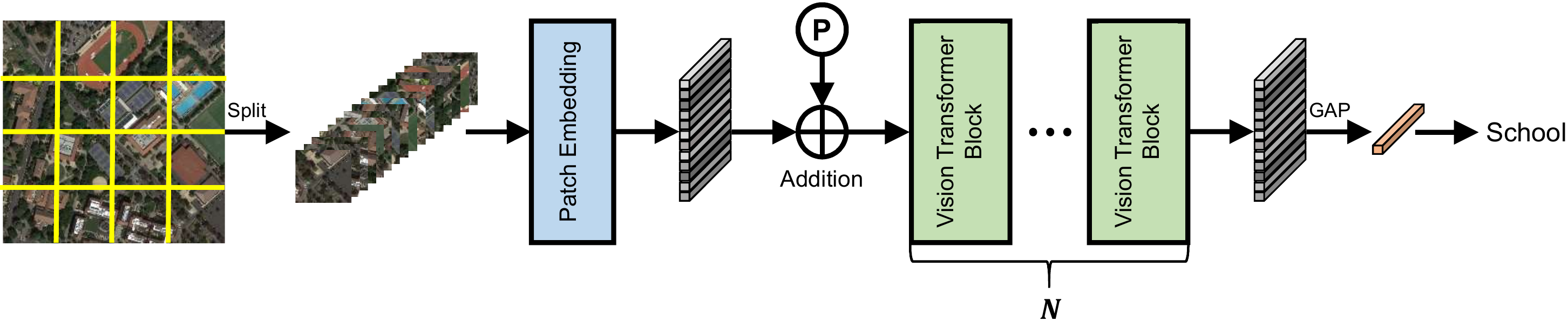}\\
  \caption{Overall structure of the pretrained vision transformer network. ``P'' means positional encoding. The final output tokens are averaged using global average pooling (GAP) for classification. 
  }
  \label{network}
\end{figure*}

\subsection{Unsupervised Pretraining by MAE}
To explore the effectiveness of MIM pretraining in RS tasks, we pretrain our model using MAE~\cite{mae} with the large-scale RSI dataset, i.e., MillionAID~\cite{Long2021DiRS}. In the following, we will briefly introduce the dataset, the MAE method, the network structure used during pretraining, and corresponding implementation details.
 
\subsubsection{MillionAID}
 
The MillionAID dataset~\cite{Long2021DiRS} is a large-scale dataset with RS scene images and corresponding labels. It contains 100,0848 non-overlapping RS scenes in RGB formats, which is suitable to serve as inputs for typical deep vision models. There are 51 classes in the dataset, where each class has about 2,000$\sim$45,000 images. MillionAID is collected from Google Earth, where images are captured with various sensors and thus have different resolutions. Generally, the image sizes in the MillionAID dataset range from 110 $\times$ 110 to 31,672 $\times$ 31,672 pixels. It should be noted that although the MillionAID dataset contains both images and labels, we only use the images during pretraining and discard the labels by following the unsupervised pretraining routine.  

\subsubsection{MAE}
MAE~\cite{mae} aims to recover the masked image parts given the visible ones with an encoder-decoder structure. Specifically, for an input image, it tokenizes the image by first splitting the image into non-overlapping patches and then projecting each patch into a visual token using a patch embedding layer. After that, several visual tokens are dropped from the inputs and treated as masked regions to be predicted, following a pre-defined mask ratio. The remained tokens are fed into the transformer encoder for feature extraction. Then, several learnable mask tokens and the extracted features of visible tokens are processed by the transformer decoder to recover the masked regions. During training, the model is optimized to minimize the distance between the recovered regions and the ground truth-masked regions, either in the pixel or feature space. We follow the original MAE setting and calculate the loss in the normalized pixel space. For more details about MAE, please refer to \cite{mae}.

\subsubsection{Pretrained Backbone Networks}

 \begin{figure}[t]
  \centering
  \includegraphics[width=0.8\linewidth]{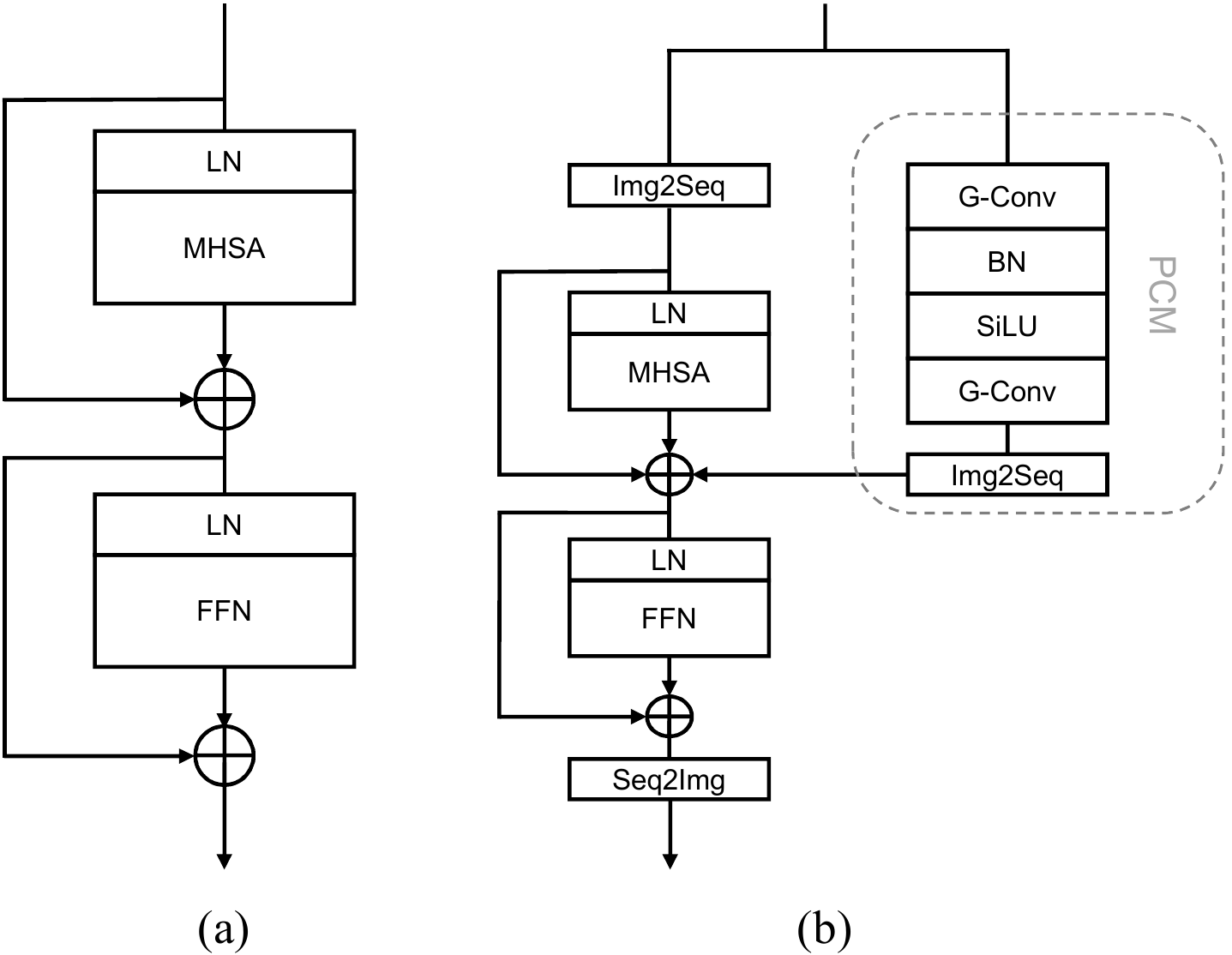}\\
  \caption{The structures of the adopted blocks in the MAE encoder. (a) ViT block. (b) Modified ViTAE normal cell. Here, BN, LN, and FFN are the batch normalization layer, layer normalization layer, and feed-forward network. G-Conv means the group convolutional layer. Img2Seq and Seq2Img are the reshape operation for conducting transformation between 1-D and 2-D features. These functions are not necessary for pretraining since there are only 1-D token sequences.
  }
  \label{transformer_block}
\end{figure}

We adopt two backbone networks for pretraining in the paper, i.e., plain ViT~\cite{vit} and ViTAE~\cite{xu2021vitae}. The former is composed of plain transformer encoders with full self-attention layers. This simple structure enables it to seamlessly work with the MAE pretraining since it discards the 2D structures and directly treats the image as a 1D sequence. By contrast, ViTAE incorporates inductive bias such as locality from convolutions along with the full self-attention layers, i.e., employing parallel convolution branches (PCM) along with the MHSA layers. It uses convolution with kernel size $1 \times 1$ in PCM during pretraining to avoid misleading inductive bias because the random masking strategy in MAE breaks the spatial relationship. Then the kernel size is padded to $3 \times 3$ when finetuning on specific downstream tasks. Assuming the weight in pretraining for $i$th convolutional layer is $\mathbf{W}_{P}^{(i)}=[\theta]_{1 \times 1}$ (ignoring the channel space), the padded kernel is implemented as follows 

 \begin{equation}
   \mathbf{W}_{F}^{(i)} = 
   \left[\begin{array}{ccc}
    \alpha & \alpha & \alpha \\
    \alpha & \theta & \alpha \\
    \alpha & \alpha & \alpha 
   \end{array}
   \right]_{3 \times 3},
 \end{equation}
where $\theta$ is the learned value during MAE and $\alpha$ is initialized to 0 and is learnable during finetuning. Besides, we adopt a shallow design of the PCM in the utilized ViTAE model, i.e., a convolutional layer, a batch normalization layer, a SiLU layer\cite{silu}, and a convolutional layer in sequence, to save the memory footprint. Figure~\ref{transformer_block} shows the plain transformer block and the basic ViTAE block used for MAE pretraining. A sin-cos positional encoding is added after the patch embedding layer to involve positional information in ViTAE (Figure~\ref{transformer_block} has not shown for simplicity). More details can be referred to~\cite{xu2021vitae, vitae_v2}.

We utilize the ``base'' version of the ViT and ViTAE models both with about 100M parameters. The networks are denoted as ``ViT-B'' and ``ViTAE-B'', respectively. The detailed structures of the two networks can be found in Table~\ref{two_networks}, where ``Patch Size'' represents the size of split patches and ``Embedding Dim'' is the dimension of the projected tokens.  ``Head'' denotes the number of heads used in MHSA. ``Group'' represents the number of groups for the convolutions in PCM, which is not used in the ViT-B model. ``Ratio'' means the expansion ratio of the FFN. ``Depth'' represents the number of transformer blocks in the two models. It has the same meaning with $N$ as demonstrated in Figure~\ref{network}.

\begin{table}[t]
  \scriptsize
  \caption{Hyper-parameter settings of ViT-B and ViTAE-B.}
  \newcommand{\tabincell}[2]{\begin{tabular}{@{}#1@{}}#2\end{tabular}}
  \centering
  \begin{tabular}{l|l|l}
  \hline
  Network& ViT-B \cite{xu2021vitae} & ViTAE-B \cite{vitae_v2} \\
  \hline
  Patch Size & 16 & 16 \\
  Embedding Dim & 768 & 768 \\
  Head & 12 & 12 \\
  Group & --- & 192 \\
  Ratio & 4 & 4 \\
  Depth & 12 & 12  \\
  \hline
\end{tabular}
  \label{two_networks}
\end{table}

\subsubsection{Implementation Details}
\label{subsubsec:implementationdetails}
We utilize two subsets $S_1$ and $S_2$ during pretraining. The two subsets have 949,848 and 51,000 images, respectively, and are randomly sampled from the MillionAID dataset. Note that a class-balance sampling strategy is employed in constructing $S_2$, i.e., we randomly select 1,000 images from each category. During pretraining, we use a batch size of 2048, equally distributed on 8 A100 GPUs, with the AdamW~\cite{adamw} optimizer. The models are trained for 1,600 epochs if not specified, following the default setting in MAE~\cite{mae}. 

\begin{table}[t]
  \scriptsize
  \caption{Results of different hyper-parameter settings in MAE for pretraining ViT-B on three datasets, i.e., UCM, AID, and NWPU. ``AVG'' denotes the average Top-1 accuracy on three datasets.}
  \newcommand{\tabincell}[2]{\begin{tabular}{@{}#1@{}}#2\end{tabular}}
  \centering
  \begin{tabular}{l|l|l|l|l}
  \hline
   Top1 Acc (\%) & UCM-55 & AID-28 & NWPU-19 & AVG \\
   \hline
   \multicolumn{5}{c}{Mask Ratio = 0.6 Epoch = 400}  \\
  \hline
  Linear probe & 45.81 & 57.51 & 59.03 & 54.12\\
  Fine tune & 99.62 & 97.22 & 94.05 & 96.96 \\
  \hline
  \multicolumn{5}{c}{Mask Ratio = 0.75 Epoch = 400}  \\
  \hline
  Linear probe & 49.90 & 61.70 & 61.28 & 57.63\\
  Fine tune & 99.71 & 97.06 & 94.43 & 97.07\\
  \hline
  \multicolumn{5}{c}{Mask Ratio = 0.9 Epoch = 400}  \\
  \hline
  Linear probe & 51.71 & 58.61 & 61.22 & 57.18 \\
  Fine tune & 99.43 & 96.33 & 93.86 & 96.54 \\
  \hline
  \multicolumn{5}{c}{Mask Ratio = 0.75 Epoch = 1600}  \\
  \hline
  Linear probe & 51.24 & 62.20 & 61.75 & 58.40 \\
  Fine tune & 99.62 & 97.53 & 94.56 & 97.24\\
  \hline
\end{tabular}
  \label{mae_mask_ratio}
\end{table}

Due to the huge distribution difference between natural images and RSIs, we first investigate the optimal mask ratio in MAE for RSIs. To this end, we search for the optimal mask ratio on the $S_1$ dataset for pretraining. We utilize a linear probing setting and finetuning setting for evaluation. We randomly initialize a linear classifier after the backbone and only update the classifier's parameters for the linear probing setting. The models are trained for 100 epochs with a batch size of 256. For the finetuning setting, parameters from the backbone network and the classifier are together tuned with a layer-wise learning rate decay mechanism~\cite{beit,mae}. We follow the same finetuning settings in MAE \cite{mae} except the batch size 64 and epoch 200. The evaluation is conducted on existing RS scene classification tasks, including UCM~\cite{ucm}, AID~\cite{aid}, and NWPU~\cite{asr_review} datasets, by finetuning the pretrained models on the three datasets. The models are pretrained for 400 epochs and finetuned for 200 epochs, respectively. The results are available in Table~\ref{mae_mask_ratio}. We report the average accuracy on the three evaluation datasets (denoted as ``AVG''). It can be observed that with the mask ratio of 0.75, the pretrained models obtain the best performance in both linear probing and finetuning settings. In the following experiments, we set the mask ratio to 0.75 by default. In addition, it can be seen that the performance could be improved when a longer training schedule is used, e.g., 1600 epochs. In the following experiments, all the pretrained models are obtained by training for 1600 epochs.

\subsection{Fintuning with Rotated Varied-Size Attention}
Different from natural images, images from RS tasks usually have larger resolutions. It will cause significant training costs in terms of floating-point operations per second (FLOPs) and memory footprint when directly transferring the pretrained models with full self-attention on the downstream tasks due to the quadratic computational complexity of full self-attention. To this end, we substitute the full self-attention modules with window-based attention modules during the finetuning stage, which reduces the computational complexity to linear with respect to the image size. This substitution is also seamless since the difference between the two kinds of attention lies in the manner of attention calculation, which is parameter-free. However, the original window-based attention operations always take fixed-size window partitions at a fixed orientation. Due to the objects of arbitrary orientations and various sizes in the RSIs, it is probably not the optimal design to use fixed-size window partitions. To address this issue, we design the rotated varied-size window attention mechanism, which will be described in detail as follows.

\subsubsection{Window-based attention and Varied-size window attention}

\noindent\textbf{Window-based attention} We will briefly review the window-based attention~\cite{swint} and varied-size window attention (VSA)~\cite{zhang2022vsa} in this part. Given an input $\mathbf{X} \in \mathbb{R}^{C \times H \times W}$, window-based attention firstly partitions it into several non-overlapping windows, i.e., $\mathbf{X}_w \in \mathbb{R}^{C \times s \times s}$, where $s$ represents the window sizes. There are a total of $\frac{H}{s} \times \frac{W}{s}$ windows in the image. Then, features inside each window is transformed to $h$ triplets representing the query, key and value features: $\{\mathbf{Q}_w^{(i)}, \mathbf{K}_w^{(i)}, \mathbf{V}_w^{(i)} \}_{i=1}^{h}$ by three linear layers, respectively, where $\mathbf{Q}_w^{(i)}, \mathbf{K}_w^{(i)}, \mathbf{V}_w^{(i)} \in \mathbb{R}^{C' \times s \times s} $, $C = hC'$, and $h$ is the number of heads in the attention layers. We denote the generated features as $\{\mathbf{Q}_w^{(i,j)}, \mathbf{K}_w^{(i,j)}, \mathbf{V}_w^{(i,j)} | j=1,\cdots, \frac{HW}{s^2} \}_{i=1}^h$, where $j$ indexes the window. After that, the attention calculations are conducted inside each window, i.e.,
\begin{equation}
  \begin{split}
    \mathbf{F}_{w}^{(i,j)} &= SA(\mathbf{Q}_w^{(i,j)},\mathbf{K}_w^{(i,j)},\mathbf{V}_w^{(i,j)}) \\
  &=softmax(\frac{\mathbf{Q}_w^{(i,j)} \mathbf{K}_w^{(i,j) T}}{\sqrt{C'}}) \mathbf{V}_w^{(i,j)} 
  \end{split},
\end{equation}
where $\mathbf{F}_{w}^{(i,j)} \in \mathbb{R}^{s^2 \times C'}$. Then, the features from different heads are concatenated along the channel dimension and features from different windows are concatenated along the spatial dimension to recover the shape of the input feature.

\noindent\textbf{Varied-size window attention} However, such a fixed-size window partition restricts the region from capturing abundant contextual information outside the handcrafted window and thus makes the model hard to handle objects of various sizes. Recently, VSA is proposed to mitigate this issue by learning the window size in a data-driven manner. Specifically, it treats the fixed-size windows as initialization and extracts the query features from these windows, i.e., $\mathbf{Q}_w$. For key and value features, the input $\mathbf{X}_w$ are used to estimate an offset and scale of the target window
\begin{equation}
  S_w, O_w = Linear(LeakyReLU(GAP(\mathbf{X}_w))),
  \label{factors}
\end{equation}
where $S_w$ and $O_w$ represent the scaling factor and the offset of the target window, and $GAP$ is short for the global average pooling operation. Then, the initial window is transformed according to the estimated two factors, i.e.,
\begin{equation}
  \left[\begin{array}{c}
    x_l \\
    y_l \\
    x_r \\
    y_r \\
  \end{array}\right]
  = \left[\begin{array}{c}
    x^c \\
    y^c \\
    x^c \\
    y^c \\
  \end{array}\right]
  + \left[\begin{array}{c}
    x_l^r \\
    y_l^r \\
    x_r^r \\
    y_r^r \\
  \end{array}\right],
\label{equ:initialLoc}
\end{equation}
\begin{equation}
  \left[\begin{array}{c}
    x_l' \\
    y_l' \\
    x_r' \\
    y_r' \\
  \end{array}\right]
  = \left[\begin{array}{c}
    x^c \\
    y^c \\
    x^c \\
    y^c \\
  \end{array}\right]
  + \left[\begin{array}{c}
    o_x \\
    o_y \\
    o_x \\
    o_y \\
  \end{array}\right]
  + \left[\begin{array}{c}
    x_l^r \cdot s_x \\
    y_l^r \cdot s_y \\
    x_r^r \cdot s_x \\
    y_r^r \cdot s_y \\
  \end{array}\right] ,
\label{equ:transformedLoc}
\end{equation}
where $x_l, y_l, x_r, y_r$ represent the coordinates of the upper left and lower right corners of the initial window. $x_c, y_c$ represent the center coordinate of the window, and $x_l^r, y_l^r, x_r^r, y_r^r$ are the distance between the corner points and the center in horizontal and vertical directions, respectively. $o_x, o_y$ and $s_x, s_y$ denote the predicted offsets and scale factors, i.e., $S_w = \{s_x, s_y \in \mathbb{R}^1 \}, O_w = \{o_x, o_y \in \mathbb{R}^1\}$. $x_l', y_l', x_r', y_r'$ indicate the corners of the transformed window. Then, the key and value features are sampled from the transformed windows and used for attention calculation. The number of sampled key and value tokens is the same as the query tokens to maintain the same computational complexity between the VSA and window-based attention.

\begin{figure}[t]
  \centering
  \includegraphics[width=1\linewidth]{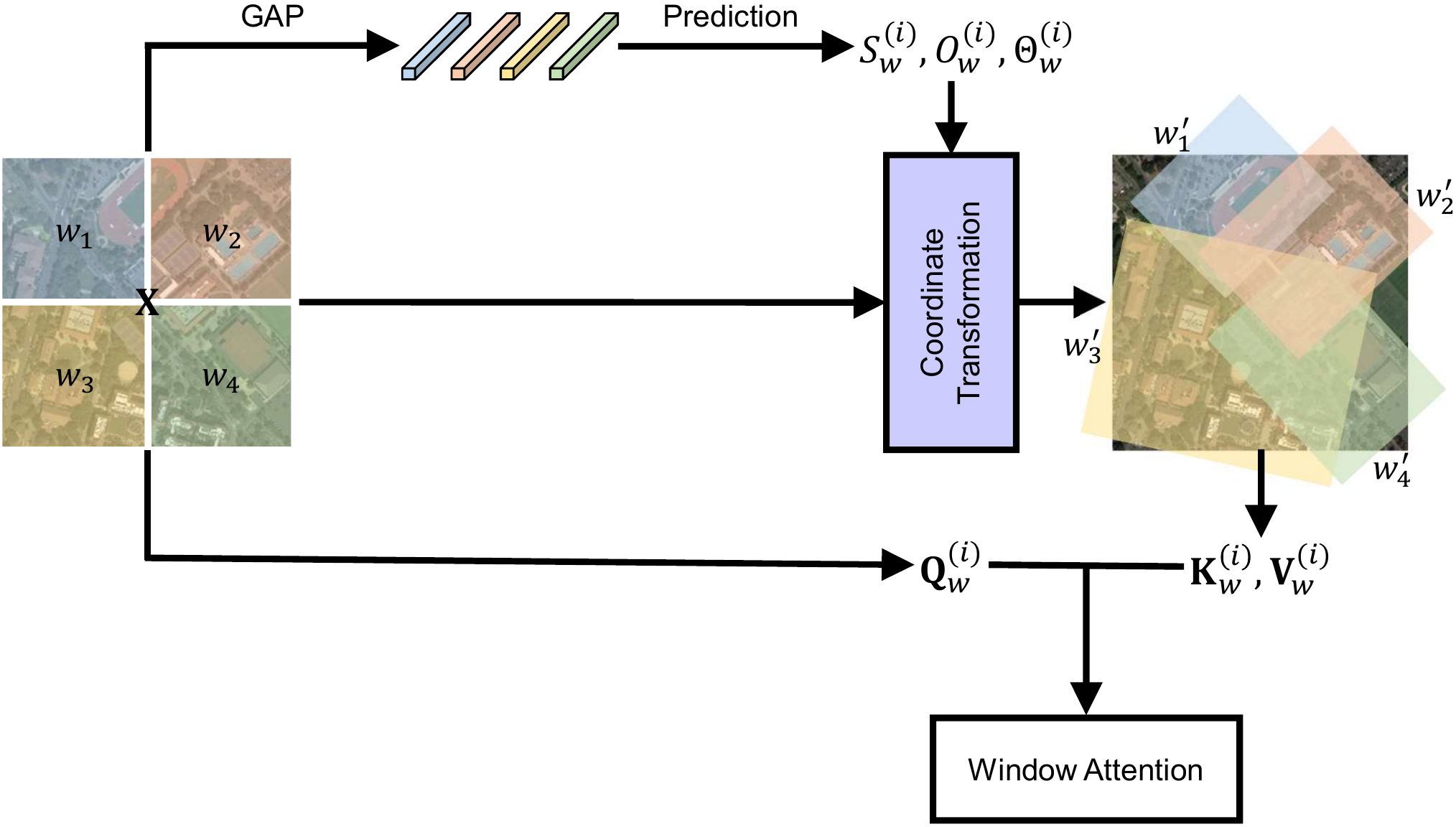}\\
  \caption{The pipeline of the proposed RVSA method in the $i$-th attention head. $w_{*}^{'}$ denotes the predicted windows for sampling key and value tokens.
  }
  \label{rvsa}
\end{figure}

\subsubsection{Rotated Varied-Size Attention} 
VSA successfully demonstrates its effectiveness in computer vision tasks on natural images. However, there are significant differences between natural images and RSIs. For example, objects in RSIs present at arbitrary orientations in images, while the default windows and the windows generated by VSA are always in the horizontal and vertical directions, which may be not suitable for RS images. To this end, we introduce one extra dimension to control the orientation of the windows, leading to the rotated varied size attention technique in this paper. Specifically, the rotation angle $\Theta_w = \{\theta \in \mathbb{R}^1\}$ is predicted along with $S_w$ and $O_w$, given the input feature $\mathbf{X}_w$, i.e.,
\begin{equation}
  S_w, O_w, \Theta_w = Linear(LeakyReLU(GAP(\mathbf{X}_w))),
  \label{three_factors}
\end{equation}
and the transformed coordinates are calculated as:
\begin{equation}
  \begin{split}
    \left[\begin{array}{c}
      x_{l/r}' \\
      y_{l/r}' \\
    \end{array}\right]
    & = \left[\begin{array}{c}
      x^c \\
      y^c \\
    \end{array}\right]
    + \left[\begin{array}{c}
      o_x \\
      o_y \\
    \end{array}\right] \\
    & + \left[\begin{array}{cc}
      \cos{\theta} & \sin{\theta} \\
      -\sin{\theta} & \cos{\theta} \\
    \end{array}\right] 
    \left[\begin{array}{c}
      x_{l/r}^r \cdot s_x\\
      y_{l/r}^r \cdot s_y\\
    \end{array}\right].
  \end{split}
\end{equation}
Figure~\ref{rvsa} shows a diagram of the proposed RVSA method.

We also propose a variant of RVSA, which allows key and value tokens can be samples from different windows, i.e., we use separate prediction layers to predict the scale, shift, and rotation factors for key and values tokens, respectively:
\begin{equation}
  \begin{split}
    S_w^K, O_w^K, \Theta_w^K  = Linear_K (LeakyReLU(GAP(\mathbf{X}_w))),\\
    S_w^V, O_w^V, \Theta_w^V  = Linear_V (LeakyReLU(GAP(\mathbf{X}_w))).
  \end{split}
\end{equation}
This more flexible module is denoted as RVSA$^\Diamond$. 

\begin{figure}[t]
  \centering
  \includegraphics[width=0.8\linewidth]{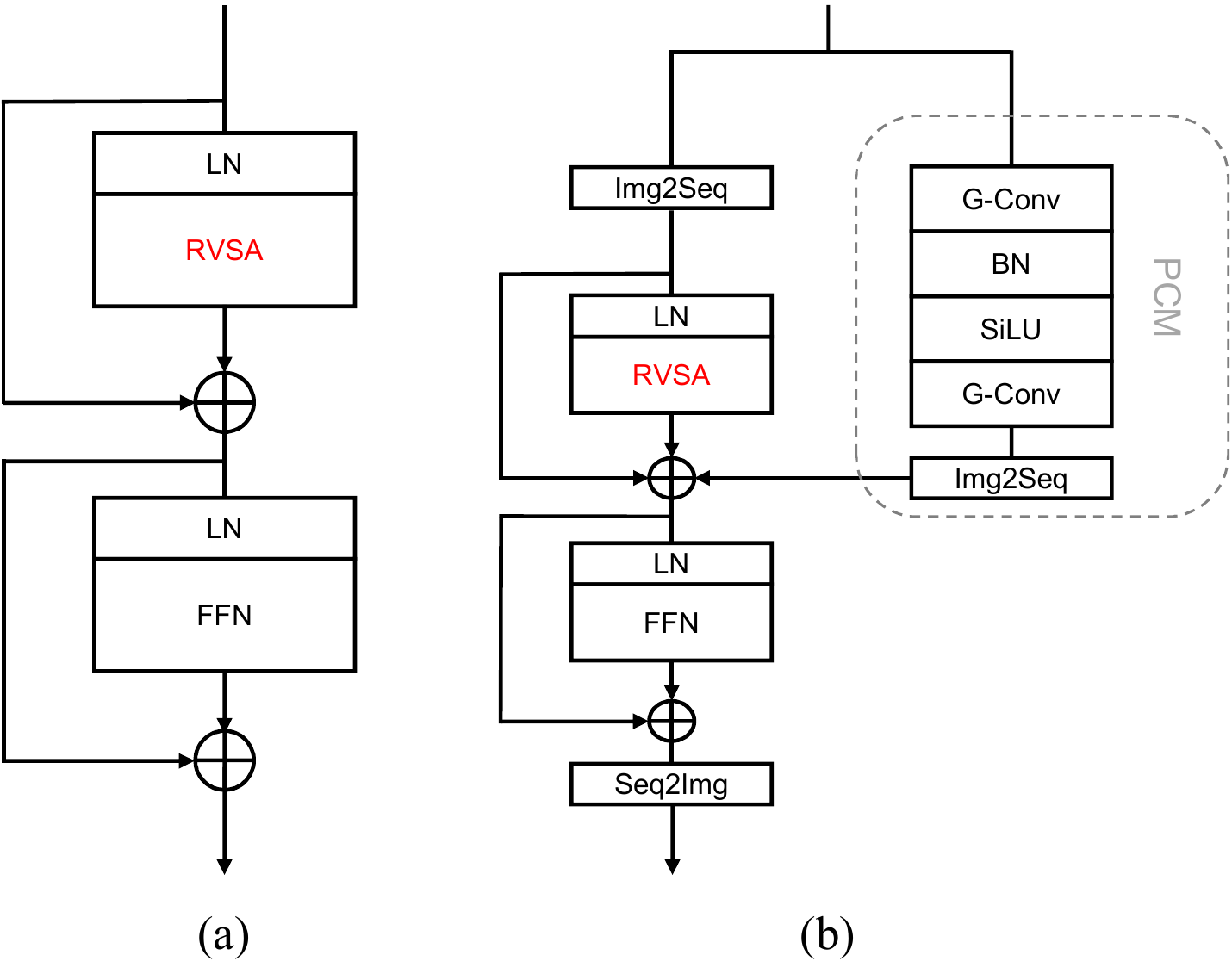}\\
  \caption{The structures of the modified blocks in (a) ViT-B + RVSA. (b) ViTAE-B + RVSA. Compare with the blocks in ViT-B and ViTAE-B, ViT-B + RVSA and ViTAE-B + RVSA simply replace MHSA with RVSA.
  }
  \label{rvsa_transformer_block}
\end{figure}

\begin{figure*}[t]
  \centering
  \includegraphics[width=0.9\linewidth]{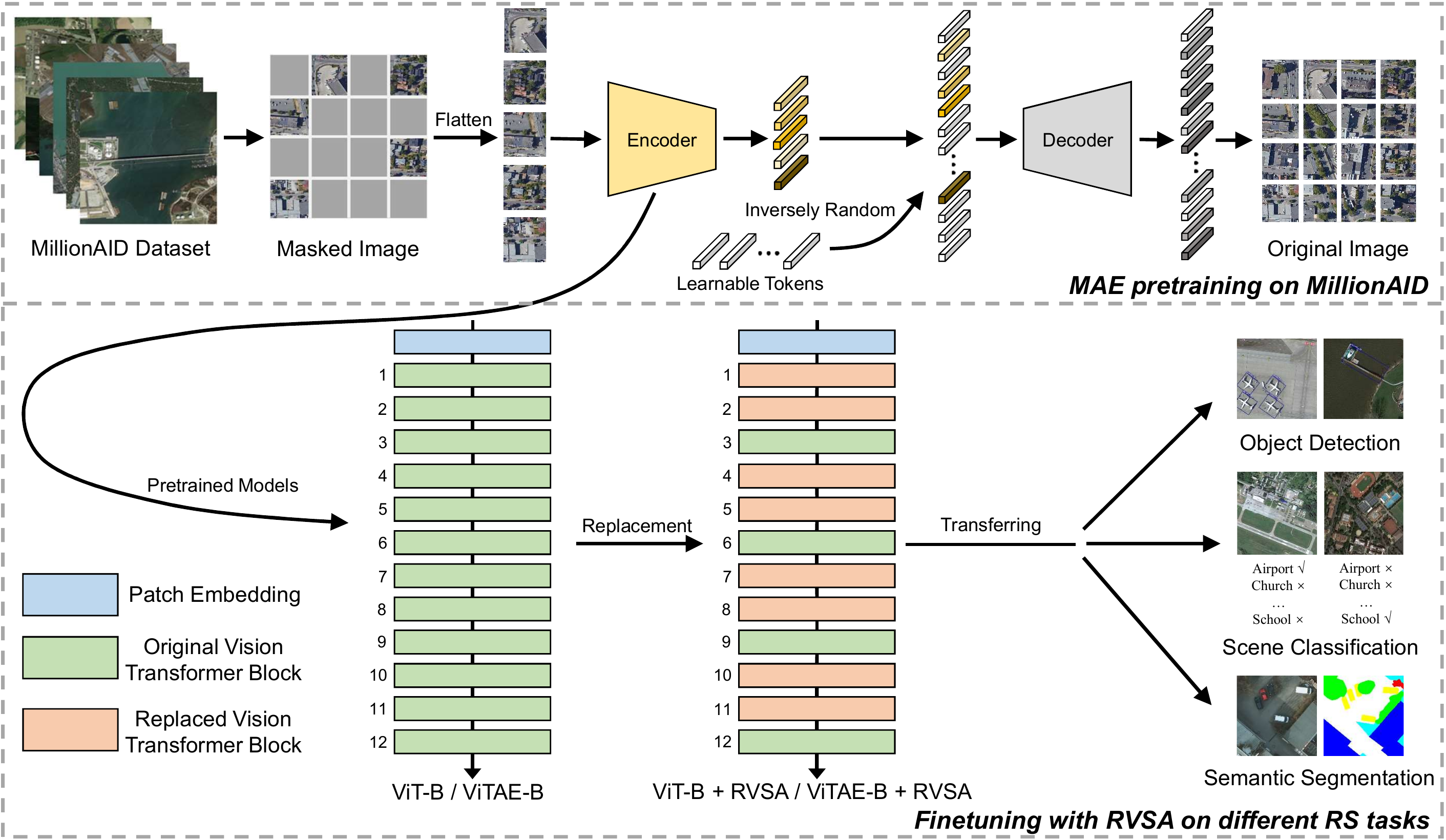}\\
  \caption{The pipeline of pretraining and finetuning. The plain ViTs are unsupervised pretrained by MAE on the large-scale RS dataset MillionAID through reconstructing the masked image patches. Then, we replace the MHSAs of some blocks in the pretrained models with different window attentions to form corresponding networks (taking RVSA as an example). The obtained networks are transferred and finetuned on different RS tasks.
  }
  \label{pipeline}
\end{figure*}

\subsubsection{Computational Complexity Analysis} We analyze the computational complexity of the proposed RVSA module. Given the input $\mathbf{X} \in \mathbb{R}^{C \times H \times W}$, we first partition the input into non-overlapping windows with the shape $s \times s$. Since complexity is the same for the calculation of each window, we will demonstrate the computational complexity of one window in this section. The features inside the window are used to predict the window's scale, shift, and rotation factors with a subsequent global average pooling (GAP) layer, an activation layer, and a linear layer. The GAP layer brings a computational complexity of $\mathcal{O}(s^2C)$. The activation layers bring about zero computations and we dismiss them for simplicity. The following linear layer projects the pooled features from dimension $C$ to dimension $5h$, it needs to predict two scale and shift factors in horizontal and vertical directions, and one rotation factor for each head. Thus, the linear projection layer has $\mathcal{O}(5hC)$ computational complexity. After that, bilinear sampling is used to sample the key and value tokens from the transformed windows, which has about $\mathcal{O}(4s^2C)$ computational complexity. Thus, the total computational complexity of RVSA for each window is $\mathcal{O}(5s^2C+5hC)$. Since there are a total number of $\frac{HW}{s^2}$ windows after the window partition, the overall extra computational complexity brought by RVSA is $\mathcal{O}(5HWC(1+\frac{5h}{s^2}))$. The computational complexity of the original window attention is $\mathcal{O}(2s^2HWC)$, where $s$ and $h$ are always set to 7 and 12. Thus, RVSA only brings marginally extra (i.e., about 11\%) computational costs.

\subsubsection{Implementation Details}
To adapt the MAE pretrained models to downstream RS tasks, we replace the plain MHSA modules with the proposed RVSA modules. Following the strategy in ViTDet~\cite{vitdet}, we use full attention blocks at each 1/4 depth layer and RVSA in all other layers. Specifically, for models with 12 layers like the ViT-B and ViTAE-B, we use full attention layers at the 3rd, 6th, 9th, and 12th layers and adopt RVSA in all other layers. The modified networks are denoted as ``ViT-B + RVSA'' and ``ViTAE-B + RVSA'' in the paper, respectively. Figure \ref{rvsa_transformer_block} shows the structure of the changed block in ViT-B + RVSA and ViTAE-B + RVSA to compare with the original ViT-B and ViTAE-B (see Figure \ref{transformer_block}). Similarly, we use the original window attention, VSA, and RVSA$^\Diamond$ for comparison. These variants are denoted as ``ViT-B-Win'', ``ViT-B + VSA'', ``ViT-B + RVSA$^\Diamond$'', ``ViTAE-B-Win'', ``ViTAE-B + VSA'', and ``ViTAE-B + RVSA$^\Diamond$'', respectively.

At last, we present the whole framework of the aforementioned pretraining and finetuning procedures in Figure \ref{pipeline} to make it easy to understand the proposed method.

\section{Experimental Results}
In this section, we evaluate the performance of the proposed models on multiple RS tasks, including scene classification, object detection, and semantic segmentation. We first conduct a series of ablation studies on the object detection task to analyze and find suitable settings for the proposed models. Then, with the help of existing popular frameworks, we compare our methods with the current SOTA approaches on public benchmarks. We also show the superiority of the proposed models in terms of computational complexity (e.g., inference speed and memory footprint) and data efficiency in transferring. 

\subsection{Remote Sensing Object Detection}
\subsubsection{Dataset}
We adopt two challenging large-scale RS oriented bounding box (OBB) detection datasets: DOTA-V1.0 \cite{dota1} and DIOR-R \cite{aod_2022_tgrs_dior_r_aopg} for evaluation.

\begin{itemize}
  \item DOTA-V1.0: It is the most famous large-scale dataset for OBB detection. It totally contains 2,806 images whose size ranges from 800 $\times$ 800 to 4,000 $\times$ 4,000, where 188,282 instances belonging to 15 categories are annotated. The training, validation, and testing sets have 1,411, 458, and 937 images, respectively.
  \item DIOR-R: It is a recently established OBB detection dataset by extending the DIOR \cite{dior}. It contains 23,463 images with a total of 192,518 instances. The trainval set and testing set contain 11,725 images with 68,073 instances, and 11,738 images with 124,445 instances, respectively. All images have been cropped to 800 $\times$ 800, where the pixel resolution ranges from 0.5 to 30m. It contains 20 common object categories.
\end{itemize}

\subsubsection{Implementation Details and Experimental Settings}
Following the \textit{de facto} standard, all models are trained with the AdamW \cite{adamw} optimizer, where the learning rate and weight decay are set to $10^{-4}$ and 0.05, respectively. We use the 1$\times$ training schedule with 12 epochs and a batch size of 2. The learning rate is adjusted by a multi-step scheduler, i.e., the learning rate is reduced by 10$\times$ at the 8th and 11th epoch. We also adopt the layer-wise learning rate decay strategy, where the decay rate sets at 0.75.
 
For a fair comparison with \cite{wang_rsp_2022}, we employ the same Oriented R-CNN \cite{orcn} detection framework for OBB detection while only changing the backbone. We adopt the default hyper-parameters defined in OBBDetection\footnote{https://github.com/jbwang1997/OBBDetection}. Following ViTDet \cite{vitdet}, we separately upsample and downsample the output feature from the last layer through deconvolution and pooling layers to construct the feature pyramid.
 
Following Oriented R-CNN, when conducting the single-scale training and testing, the DOTA dataset is cropped to 1,024 $\times$ 1,024 patches with a stride of 824. We also implement the multiscale training and testing, where the original images are first resized to three scales, i.e., (0.5, 1.0, 1.5), which are then cropped to 1,024 $\times$ 1,024 patches with a stride of 524. During training, we adopt the data augmentations including random horizontal and vertical flipping, while the random rotation is considered for multiscale training and testing. For DOTA-V1.0, the original training and validation sets are jointly used for training following~\cite{wang_rsp_2022}. We separately evaluate our models on the original testing sets of DOTA-V1.0 and DIOR-R. The evaluation results of DOTA-V1.0 are obtained from the online server by submitting the predictions of the testing set. The average precision (AP) of each class and the mean average precision (mAP) are reported. All models are trained using NVIDIA A100 GPUs.

\subsubsection{Determining the Suitable Window Size}
\begin{table}[t]
  \scriptsize
  \caption{The mAP (\%) of ViT-B + RVSA with different window sizes on DOTA-V1.0 and DIOR-R datasets.}
  \newcommand{\tabincell}[2]{\begin{tabular}{@{}#1@{}}#2\end{tabular}}
  \centering
  \begin{tabular}{l|cccc}
  \hline
  Window size  & 4 & 7 & 11 & 14 \\
  \hline
  DOTA-V1.0 & 77.84 & \bfseries 78.75 & 77.83 & 77.44 \\
  DIOR-R & 70.55 &  \bfseries 70.67 & 70.40 & 70.17 \\
  \hline
\end{tabular}
  \label{window_size}
\end{table}

Considering ViTDet \cite{vitdet} and VSA \cite{zhang2022vsa} use different window sizes, we first investigate the influence of window size $s$ on DOTA-V1.0 and DIOR-R. We search for the optimal value of $s$ in the range of $[4, 14]$ and the results are shown in Table \ref{window_size}. Interestingly, the performance peaks at $s=7$ on both datasets, and either increasing or decreasing the window size will lead to a performance drop. The reason can be attributed to two aspects: (1) Although a larger window size increases the receptive field and makes the tokens within each window encode more context, it limits the window diversity due to the reduced window number $\frac{HW}{s^2}$ and thus leads to less-diverse contexts extracted by the model. (2) When decreasing the window size, the number of the attended tokens will also drop, resulting in inefficient information captured during attention. The experiments indicate that $s=7$ can be a good balance between the token number and window diversity for these RS datasets. Thus we set $s=7$ by default in the following experiments.

\begin{table}[t]
  \caption{The mAP (\%) of different ViT-B variants on DOTA-V1.0 and DIOR-R datasets. The suffix denotes different window attention methods. WA: window attention. SF: scale factor. OF: offset factor. RF: rotation factor.}
  \newcommand{\tabincell}[2]{\begin{tabular}{@{}#1@{}}#2\end{tabular}}
  \centering
  \begin{threeparttable}
  \resizebox{\linewidth}{!}{
  \begin{tabular}{l|ccc|cc}
  \hline
  \multirow{2}*{Method} & \multirow{2}*{WA} & \multirow{2}*{SF \& OF}  & \multirow{2}*{RF} & \multicolumn{2}{c}{Oriented R-CNN mAP}\\
  & & & & DOTA-V1.0 & DIOR-R \\
  \hline
  ViT-B &  &  &  & 77.05  & 66.65 \\
  ViT-B-Win & \ding{52} &  & & 77.19 & 67.95 \\
  ViT-B + VSA & \ding{52} &  \ding{52} & & 78.40 & 70.48  \\
  ViT-B + RVSA & \ding{52} &  \ding{52}& \ding{52} & \bfseries 78.75 & 70.67 \\
  ViT-B + RVSA$^\Diamond$ & \ding{52} & \ding{52} & \ding{52} & 78.61 & \bfseries 70.85  \\
  \hline
\end{tabular}
  }
  \end{threeparttable}
  \label{vit_windows}
\end{table}

\begin{table}[t]
  \caption{The training costs of different ViT-B variants on DOTA-V1.0 and DIOR-R datasets.}
  \newcommand{\tabincell}[2]{\begin{tabular}{@{}#1@{}}#2\end{tabular}}
  \centering
  \begin{threeparttable}
  \resizebox{\linewidth}{!}{
  \begin{tabular}{l|ccccc}
  \hline
  \multirow{2}*{Method} & \multicolumn{5}{c}{Oriented R-CNN} \\
  \cline{2-6}
   & Params (M)  & Input Size & FLOPs (G)\tnote{1} & Memory (M) & $T_{trn}$ (hh:mm:ss) \\
  \hline
  \bfseries \textit{DOTA-V1.0} & \multicolumn{5}{c}{}\\
  \hline
  ViT-B\tnote{2} & 113.73 & 1,024 & 717.79 & 25,757*2 & 12:30:42 \\
  ViT-B-Win & 113.63  & 1,024  & 427.43 & 24,685 & 11:41:29 \\ 
  ViT-B + VSA & 113.92  & 1,024 & 413.26  & 25,321 & 12:12:11 \\ 
  ViT-B + RVSA & 114.00  & 1,024 & 413.29 & 25,343 & 12:31:30 \\
  ViT-B + RVSA$^\Diamond$ & 114.37 &  1,024  & 413.60 & 25,386 & 13:08:39 \\
  \hline
  \bfseries \textit{DIOR-R} & \multicolumn{5}{c}{}\\
  \hline
  ViT-B & 112.47 &   800 & 364.56 & 21,408 & 09:05:08 \\
  ViT-B-Win & 112.39 & 800 & 263.63 & 12,322 & 06:12:14 \\
  ViT-B + VSA & 112.69 &  800 & 252.37 & 12,729 & 06:56:16 \\
  ViT-B + RVSA & 112.76 & 800 &  252.40 & 12,744 & 07:11:35 \\
  ViT-B + RVSA$^\Diamond$ & 113.13 &  800 & 252.48 & 12,772 & 07:43:18 \\
  \hline
\end{tabular}
  }
  \begin{tablenotes}
    \scriptsize
    \item[1] The FLOPs are calculated for each backbone network.
    \item[2] This model is trained on 2 GPUs since it encounters the out-of-memory \\ issue on a single GPU, while other models are trained on a single GPU.
  \end{tablenotes}
  \end{threeparttable}
  \label{complexity}
\end{table}

\subsubsection{Comparison of Different Attention Methods}

\begin{figure*}[t]
  \centering
  \includegraphics[width=0.8\linewidth]{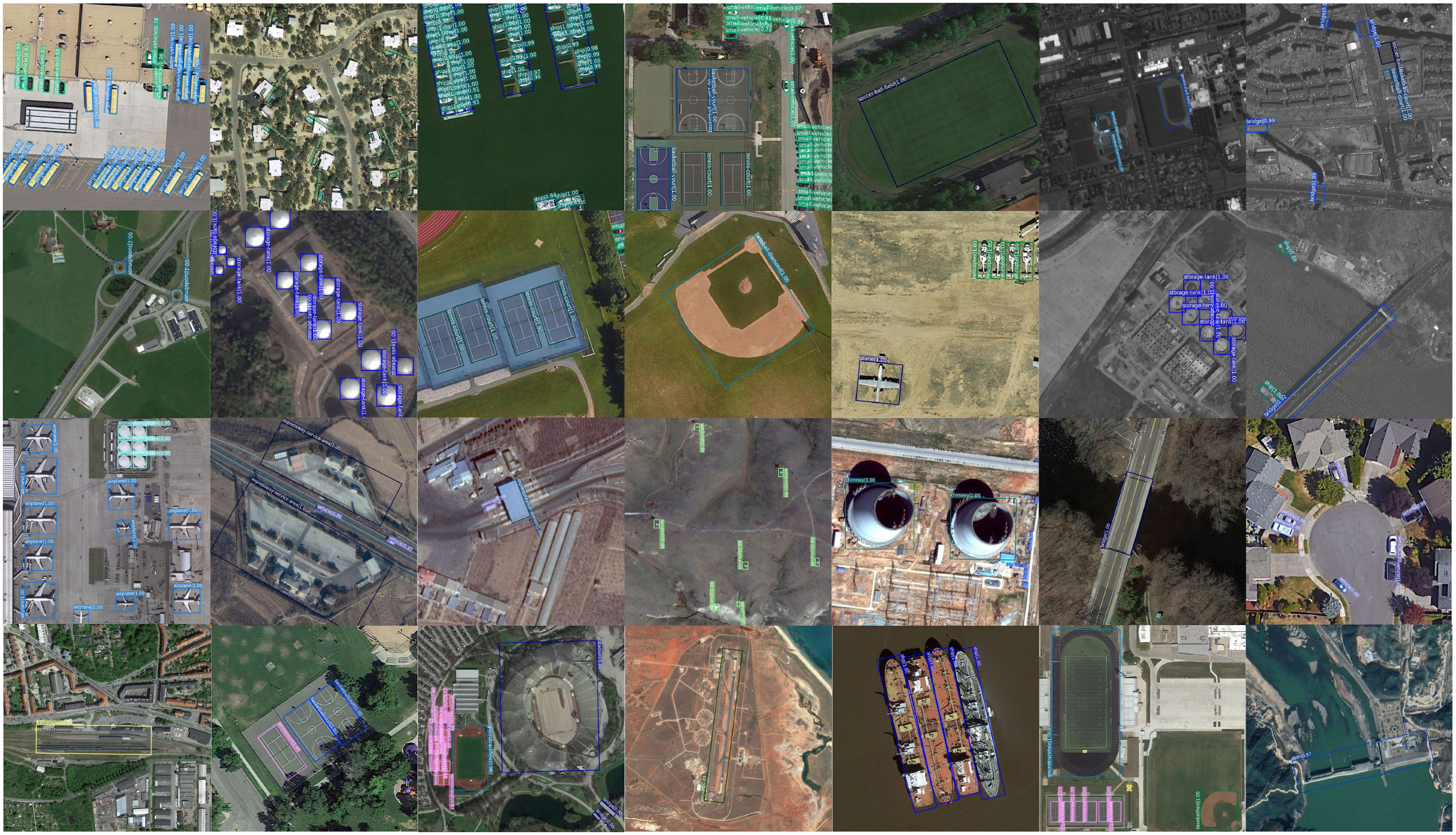}\\
  \caption{Some visual detection results. The first two rows are the results of ViTAE-B + RVSA on DOTA-V1.0, while the remained rows are the results of ViTAE-B + RVSA$^\Diamond$ on DIOR-R. Best viewed with zoom-in. 
  }
  \label{det_vis}
\end{figure*}

\label{subsubsec:attentioncomparison}
After determining the window size $s$, we compare different attention methods based on ViT-B and the results are listed in Table \ref{vit_windows}. Surprisingly, ViT-B with full self-attention performs worse than ViT-B-Win despite more tokens involved during attention. By introducing the scale and offset factors, the size and location of the window in VSA can be adaptively learned. Compared with the ViT-B-Win, ViT-B + VSA learns adaptive windows for varied-size objects in different positions, improving the detection performance significantly. Nevertheless, the learned windows are horizontal and vertical, which are insufficient to accurately capture the object context in the RSI that may be displayed in any direction. ViT-B + RVSA addresses this issue by introducing an extra rotation factor, thus generating windows in various directions to better fit the overhead viewing characteristics in RS. It enables the model to effectively extract more suitable contexts for describing RS objects. As a result, ViT-B + RVSA performs better than ViT-B + VSA on both DOTA-V1.0 and DIOR-R. At last, we evaluate ViT-B + RVSA$^\Diamond$ where the key and value tokens are sampled from different windows to extract contexts more flexibly. Table \ref{vit_windows} shows that ViT-B + RVSA and ViT-B + RVSA$^\Diamond$ perform slightly different on the two datasets. We guess that relaxing the constraint that key and value tokens share the same window can further improve the model representation capability while at the risk of overfitting. Therefore, ViT-B + RVSA$^\Diamond$ performs better on the more challenging dataset DIOR-R, which has more training images and categories.

Besides accuracy, we also compare the training cost of the above models. Table \ref{complexity} summarizes the evaluation metrics including the number of parameters (Params), computations (FLOPs), GPU memory footprint, and training time (shown as $T_{trn}$). It can be seen that all these models have more than 100M parameters, where ViT-B has the largest memory footprint and FLOPs as well as the longest training time (Note that it uses two GPUs) because of the quadratic complexity of full self-attention. ViT-B-Win alleviates these issues by adopting WMHSA, where the parameters reduce slightly because of the use of relative positional encoding instead of absolute positional encoding. Note that the FLOPs of ViT-B + VSA are smaller than ViT-B-Win since the padding operation is implemented after the generation of query, key, and value tokens. ViT-B + VSA brings slightly extra memory footprints than ViT-B-Win due to learnable scale and offset factors. Compared to ViT-B + VSA, ViT-B + RVSA has a similar complexity, while ViT-B + RVSA$^\Diamond$ slightly increases the parameters and computational overheads since it adopts individual window prediction layers for key and value tokens. Compared to ViT-B, the proposed ViT-B + RVSA and ViT-B + RVSA$^\Diamond$ can save approximately half the memory and accelerate the training speed, while achieving better performance.

\begin{table*}[ht]
  \scriptsize
  \caption{Results of different methods on the testing set of the DOTA-V1.0 dataset.}
  \newcommand{\tabincell}[2]{\begin{tabular}{@{}#1@{}}#2\end{tabular}}
  \centering
  \begin{threeparttable}
  \resizebox{\linewidth}{!}{
  \begin{tabular}{l|l|l|ccccccccccccccc|c}
  \hline
  Method & Pretrain &Backbone & PL\tnote{1} & BD & BR & GTF & SV & LV & SH & TC & BC & ST & SBF & RA & HA & SP & HC & mAP \\
  \hline
  \bfseries \textit{Singe-scale} & \multicolumn{17}{c}{}\\
  \hline
  ROI Trans. \cite{roi_transformer} & IMP\tnote{2} & ResNet-101 & 88.64 & 78.52 & 43.44 & \textbf{\textcolor{blue}{75.92}} & 68.81 & 73.68 & 83.59 & 90.74 & 77.27 & 81.46 & 58.39 & 53.54 & 62.83 & 58.93 & 47.67 & 69.56 \\
  R$^3$Det \cite{r3det} & IMP & ResNet-101 & 88.76 & 83.09 & 50.91 & 67.27 & 76.23 & 80.39 & 86.72 & 90.78 & 84.68 & 83.24 & 61.98 & 61.35 & 66.91 & 70.63 & 53.94 & 73.79 \\
  Gilding Vertex \cite{glid_vertex} & IMP & ResNet-50 & \textbf{\textcolor{red}{89.64}} & \bfseries 85.00& 52.26 & \bfseries 77.34 & 73.01 & 73.14 & 86.82 & 90.74 & 79.02 & \textbf{\textcolor{red}{86.81}} & 59.55 & 70.91 & 72.94 & 70.86& 57.32 & 75.02 \\
  AOPG \cite{aod_2022_tgrs_dior_r_aopg} & IMP& ResNet-101 & 89.14 & 82.74 & 51.87 & 69.28 & 77.65 & 82.42 & 88.08 & \textbf{\textcolor{red}{90.89}} & 86.26 & 85.13 & 60.60 & 66.30 & 74.05 & 67.76 & 58.77 & 75.39 \\
  DODet \cite{aod_2022_tgrs_dodet} &IMP & ResNet-101 & \textbf{\textcolor{blue}{89.61}} & 83.10 & 51.43 & 72.02 & 79.16 & 81.99 & 87.71 & \textbf{\textcolor{red}{90.89}} & 86.53 & 84.56 & 62.21 & 65.38 & 71.98 & 70.79 & 61.93 & 75.89 \\
  S$^2$ANet \cite{aod_2022_tgrs_s2anet} &IMP & ResNet-101 & 88.70 & 81.41 & 54.28 & 69.75 & 78.04 & 80.54 & 88.04 & 90.69 & 84.75 & 86.22 & \textbf{\textcolor{red}{65.03}} & 65.81 & 76.16 & 73.37& 58.86 & 76.11 \\
  ReDet \cite{aod_2021_cvpr_redet} & IMP & ReResNet-50 & 88.79 & 82.64 & 53.97 & 74.00 & 78.13 & 84.06 & 88.04 & \textbf{\textcolor{red}{90.89}} & 87.78 & 85.75 & 61.76 & 60.39 & 75.96 & 68.07 & 63.59 & 76.25 \\
  R$^3$Det-KLD \cite{aod_2021_nips_kld} & IMP & ResNet-50 & 88.90 & 84.17 & 55.80 & 69.35 & 78.72 & 84.08 & 87.00 & 89.75 & 84.32 & 85.73 & \textbf{\textcolor{blue}{64.74}} & 61.80 & 76.62 & 78.49 & 70.89 & 77.36 \\
  Oriented RepPoints \cite{aod_2022_cvpr_oriented_reppoint} & IMP & Swin-T & 89.11 & 82.32 & 56.71 & 74.95 & \textbf{\textcolor{red}{80.70}} & 83.73 & 87.67 & 90.81 & 87.11 & 85.85 & 63.60 & 68.60 & 75.95 & 73.54 & 63.76 & 77.63 \\
  Oriented R-CNN \cite{orcn} &IMP & ResNet-101 & 88.86 & 83.48 & 55.27 & \textbf{\textcolor{red}{76.92}} & 74.27 & 82.10 & 87.52 & \bfseries 90.90 & 85.56 & 85.33 & \bfseries 65.51 & 66.82 & 74.36 & 70.15 & 57.28 & 76.28 \\
  Oriented R-CNN \cite{orcn}& RSP \cite{wang_rsp_2022} & ViTAEv2-S & \bfseries 89.66 & 83.04 & 55.85 & 75.16 & 79.95 & 84.34 & 88.04 & \bfseries 90.90 & 88.17 & 85.58 & 62.64 & 70.60 & 76.77 & 67.15 & 67.89 & 77.72 \\
  AO2-DETR \cite{aod_2022_ao2detr} &IMP & ResNet-50 & 89.27 & \textbf{\textcolor{red}{84.97}} & 56.67 & 74.89 & 78.87 & 82.73 & 87.35 & 90.50 & 84.68 & 85.41 & 61.97 & 69.96 & 74.68 & 72.39 & \textbf{\textcolor{blue}{71.62}} & 77.73 \\
  \hline
  Oriented R-CNN & MAE & ViT-B & 89.33 & 79.82 & 54.99 & 70.83 & 79.89 & 85.25 & 88.01 & \textbf{\textcolor{blue}{90.88}} & 83.83 & 86.23 & 56.30 & 68.19 & 75.98 & 76.42 & 69.93 & 77.05 \\
  Oriented R-CNN & MAE & ViTAE-B & 89.40 & 80.02 & 55.25 & 70.90 & \bfseries 80.84 & 85.45 & 88.20 & 90.87 & 83.56 & 86.39 & 56.54 & 69.87 & 77.11 & \textbf{\textcolor{red}{79.63}} & 67.47 & 77.43 \\
  Oriented R-CNN & MAE & ViT-B + VSA & 89.19 & 83.32 & 58.10 & 70.91 & 79.73 & \bfseries 85.87 & \textbf{\textcolor{blue}{88.34}} & \bfseries 90.90 & \bfseries 88.87 & 86.58 & 57.61 & \textbf{\textcolor{red}{72.31}} & \bfseries 77.35 & 78.46 & 68.45 & 78.40 \\
  Oriented R-CNN & MAE & ViTAE-B + VSA &89.55 & 81.98 & 58.38 & 71.43 & \textbf{\textcolor{blue}{80.27}} & 85.24 & \textbf{\textcolor{red}{88.36}} & 90.87 & 88.35 & 86.32 & 59.26 & 71.87 & \bfseries 77.35 & \bfseries 81.09 & 68.76 & 78.60 \\
  \hline
  Oriented R-CNN & MAE & ViT-B + RVSA (Ours) & 89.07 & 83.45 & \textbf{\textcolor{blue}{59.32}} & 72.15 & 80.10 &  \textbf{\textcolor{red}{85.72}} & \bfseries 88.41 & 90.85 & \textbf{\textcolor{red}{88.55}} & \bfseries 87.14 & 58.53 & 69.63 & 76.71 & 79.10 & \bfseries 72.52 & \textbf{\textcolor{blue}{78.75}} \\
  Oriented R-CNN & MAE & ViT-B + RVSA$^\Diamond$ (Ours) & 89.23 & 81.44 & 57.91 & 72.94 & 79.91 & 85.08 & 88.23 & 90.87 & 87.37 & 86.68 & 59.07 & \bfseries 73.62 & 77.12 & 78.70 & 71.06 & 78.61 \\
  Oriented R-CNN & MAE & ViTAE-B + RVSA (Ours) & 89.35 & 83.46 & \bfseries 59.97 & 72.37 & 79.79 & \textbf{\textcolor{blue}{85.58}} & 88.21 & 90.87 & 87.62 & \textbf{\textcolor{blue}{86.75}} & 60.48 & 71.63 & \textbf{\textcolor{blue}{77.26}} & 78.65 & \textbf{\textcolor{red}{72.36}} & \textbf{\textcolor{red}{78.96}} \\
  Oriented R-CNN & MAE & ViTAE-B + RVSA$^\Diamond$ (Ours) & 89.38 & \textbf{\textcolor{blue}{84.26}} & \textbf{\textcolor{red}{59.39}} & 73.19 & 79.99 & 85.36 & 88.08 & 90.87 & \textbf{\textcolor{blue}{88.50}} & 86.53 & 58.93 & \textbf{\textcolor{blue}{72.24}} & \textbf{\textcolor{red}{77.31}} & \textbf{\textcolor{blue}{79.59}} & 71.24 & \bfseries 78.99 \\
  \hline
  \bfseries \textit{Multi-scale} & \multicolumn{17}{c}{}\\
  \hline
  R$^3$Det \cite{r3det} &IMP & ResNet-152 & 89.80 & 83.77 & 48.11 & 66.77 & 78.76 & 83.27 & 87.84 & 90.82 & 85.38 & 85.51 & 65.67 & 62.68 & 67.53 & 78.56 & 72.62 & 76.47 \\
  AO2-DETR \cite{aod_2022_ao2detr} &IMP & ResNet-50 & \textbf{\textcolor{red}{89.95}} & 84.52 & 56.90 & 74.83 & \textbf{\textcolor{blue}{80.86}} & 83.47 & 88.47 & 90.87 & 86.12 & \bfseries 88.55 & 63.21 & 65.09 & 79.09 & \textbf{\textcolor{red}{82.88}} & 73.46 & 79.22 \\
  S$^2$ANet \cite{aod_2022_tgrs_s2anet} & IMP & ResNet-50 & 88.89 & 83.60 & 57.74 & 81.95 & 79.94 & 83.19 & \bfseries 89.11 & 90.78 & 84.87 & 87.81 & 70.30 & 68.25 & 78.30 & 77.01 & 69.58 & 79.42 \\
  ReDet \cite{aod_2021_cvpr_redet} & IMP& ReResNet-50 & 88.81 & 82.48 & 60.83 & 80.82 & 78.34 & \textbf{\textcolor{blue}{86.06}} & 88.31 & 90.87 & 88.77 & 87.03 & 68.65 & 66.90 & 79.26 & 79.71 & 74.67 & 80.10 \\
  R$^3$Det-GWD \cite{aod_2021_icml_gwd} & IMP & ResNet-152 & 89.66 & 84.99 & 59.26 & \textbf{\textcolor{red}{82.19}} & 78.97 & 84.83 & 87.70 & 90.21 & 86.54 & 86.85 & \bfseries 73.47 & 67.77 & 76.92 & 79.22 & 74.92 & 80.23 \\
  ReDet-DEA \cite{aod_2022_tgrs_dea} &IMP & ReResNet-50 & \textbf{\textcolor{blue}{89.92}} & 83.84 & 59.65 & 79.88 & 80.11 & \bfseries 87.96 & 88.17 & 90.31 & 88.93 & \textbf{\textcolor{red}{88.46}} & 68.93 & 65.94 & 78.04 & 79.69 & 75.78 & 80.37 \\ 
  DODet \cite{aod_2022_tgrs_dodet} & IMP& ResNet-50 & \bfseries 89.96 & 85.52 & 58.01 & 81.22 & 78.71 & 85.46 & 88.59 & \textbf{\textcolor{red}{90.89}} & 87.12 & 87.80 & 70.50 & 71.54 & 82.06 & 77.43 & 74.47 & 80.62 \\
  R$^3$Det-KLD \cite{aod_2021_nips_kld} &IMP & ResNet-152 & \textbf{\textcolor{blue}{89.92}} & 85.13 & 59.19 & 81.33 & 78.82 & 84.38 & 87.50 & 89.80 & \textbf{\textcolor{red}{87.33}} & 87.00 & \textbf{\textcolor{blue}{72.57}} & 71.35 & 77.12 & 79.34 & 78.68 & 80.63 \\
  AOPG \cite{aod_2022_tgrs_dior_r_aopg} &IMP & ResNet-50 & 89.88 & 85.57 & 60.90 & 81.51 & 78.70 & 85.29 & \textbf{\textcolor{blue}{88.85}} & \textbf{\textcolor{red}{90.89}} & \bfseries 87.60 & 87.65 & 71.66 & 68.69 & 82.31 & 77.32 & 73.10 & 80.66 \\
  Oriented R-CNN \cite{orcn} &IMP & ResNet-50 & 89.84 & 85.43 & 61.09 & 79.82 & 79.71 & 85.35 & 88.82 & \textbf{\textcolor{blue}{90.88}} & 86.68 & 87.73 & 72.21 & 70.80 & 82.42 & 78.18 & 74.11 & 80.87 \\
  ROI Trans.-KFIoU \cite{aod_2022_kfiou} &IMP & Swin-T & 89.44 & 84.41 & \bfseries 62.22 & \bfseries 82.51 & 80.10 & \textbf{\textcolor{red}{86.07}} & 88.68 & \bfseries 90.90 & \textbf{\textcolor{blue}{87.32}} & \textbf{\textcolor{blue}{88.38}} & \textbf{\textcolor{red}{72.80}} & 71.95 & 78.96 & 74.95 & 75.27 & 80.93 \\
  \hline
  Oriented R-CNN & MAE & ViT-B & 89.46 & 84.32 & \textbf{\textcolor{blue}{61.81}} & 79.74 & 79.66 & 85.54 & 88.47 & 90.85 & 82.19 & 86.58 & 62.04 & \textbf{\textcolor{blue}{72.48}} & \textbf{\textcolor{blue}{84.63}} & 80.72 & 78.43 & 80.46 \\
  Oriented R-CNN & MAE & ViTAE-B & 87.47 & 84.90 & 60.97 & 79.23 & 80.62 & 85.08 & 88.47 & 90.84  & 85.66 & 87.32 & 62.11 & 70.27 & 83.82 & \textbf{\textcolor{blue}{82.03}} &78.67 & 80.50 \\
  Oriented R-CNN & MAE & ViT-B + VSA & 88.27 & \textbf{\textcolor{red}{85.81}} & \textbf{\textcolor{red}{61.88}} & 79.75 & 79.76 & 85.07 & 88.30 & 90.87 & 84.12 & 86.67 & 62.79 & 71.95 & 84.27 & 81.85 & \bfseries 81.48 & 80.86 \\
  Oriented R-CNN & MAE & ViTAE-B + VSA & 88.95 & \bfseries 85.96 & 60.69 & 80.93 & \textbf{\textcolor{red}{80.88}} & 84.61 & 88.58 & 90.84 & 83.28 & 87.31 & 65.36 & 72.44 & 84.16 & 81.37 & \textbf{\textcolor{red}{80.02}} & \textbf{\textcolor{blue}{81.03}} \\
  \hline
  Oriented R-CNN & MAE & ViT-B + RVSA (Ours) & 87.63 & 85.23 & 61.73 & 81.11 & 80.68 & 85.37 & 88.26 & 90.80 & 86.38 & 87.21 & 67.93 & 69.81 & 84.06 & 81.25 & 77.76 & 81.01 \\
  Oriented R-CNN & MAE & ViT-B + RVSA$^\Diamond$ (Ours) & 87.28 & 85.40 & 59.83 & 80.80 & 80.37 & 85.49 & 88.45 & 90.87 & 85.39 & 87.05 & 64.09 & \textbf{\textcolor{red}{72.70}} & 84.15 & 81.84 & 78.30 & 80.80 \\    
  Oriented R-CNN & MAE & ViTAE-B + RVSA (Ours) & 88.97 & \textbf{\textcolor{blue}{85.76}} & 61.46 & 81.27 & 79.98 & 85.31 & 88.30 & 90.84 & 85.06 & 87.50 & 66.77 & \bfseries 73.11 &\bfseries 84.75 & 81.88 & 77.58 & \bfseries 81.24  \\
  Oriented R-CNN & MAE & ViTAE-B + RVSA$^\Diamond$ (Ours) & 89.40 & 83.94 & 59.76 & \textbf{\textcolor{blue}{82.10}} & \bfseries 81.73 & 85.32 & \textbf{\textcolor{red}{88.88}} & 90.86 & 85.69 & 87.65 & 63.70 & 69.94 & \textbf{\textcolor{red}{84.72}} & \bfseries 84.16 &\textbf{\textcolor{blue}{79.90}} & \textbf{\textcolor{red}{81.18}}  \\
  \hline
\end{tabular}
  }
  \begin{tablenotes}
    \scriptsize
    \item[1] PL: plane. BD: baseball diamond. BR: bridge. GTF: ground track field. SV: small vehicle. LV: large vehicle. SH: ship. TC: tennis court. BC: baseball court. ST: storage tank. \\ SBF: soccer ball field. RA: roundabout. HA: harbor. SP: swimming pool. HC: helicopter.
    \item[2] IMP: ImageNet pretraining. RSP: remote sensing supervised pretraining on the MillionAID. MAE: MAE unsupervised pretraining on the MillionAID. 
  \end{tablenotes}
  \end{threeparttable}
  \label{det_dota}
\end{table*}

\begin{table*}[ht]
  \scriptsize
  \caption{Results of different methods on the testing set of the DIOR-R dataset.}
  \newcommand{\tabincell}[2]{\begin{tabular}{@{}#1@{}}#2\end{tabular}}
  \centering
  \begin{threeparttable}
  \resizebox{\linewidth}{!}{
  \begin{tabular}{l|l|l|cccccccccccccccccccc|c}
  \hline
  Method & Pretrain &Backbone & APL\tnote{1} & APO & BF & BC & BR & CH & DAM & ETS & ESA & GF & GTF & HA & OP & SH & STA & STO & TC & TS & VE & WM  & mAP \\
  \hline
  RetinaNet-O \cite{focal_loss} & IMP & ResNet-50 & 61.49 & 28.52 & 73.57 & 81.17& 23.98 &72.54 & 19.94 & 72.39 & 58.20 & 69.25 & 79.54 & 32.14& 44.87 & 77.71 & 67.57 & 61.09 & 81.46 & 47.33 & 38.01 & 60.24 & 57.55 \\
  Faster R-CNN-O \cite{FasterRCNN} & IMP & ResNet-50 & 62.79 & 26.80& 71.72 & 80.91 & 34.20 & 72.57 & 18.95 & 66.45& 65.75 & 66.63 & 79.24 & 34.95 & 48.79 & 81.14 & 64.34 & 71.21 & 81.44 & 47.31 & 50.46 & 65.21& 59.54 \\ 
  ROI Trans. \cite{roi_transformer} & IMP & ResNet-50 & 63.34 & 37.88 & 71.78 & 87.53 & 40.68 & 72.60 & 26.86 & \textbf{\textcolor{red}{78.71}} & 68.09& 68.96 & 82.74 & \textbf{\textcolor{red}{47.71}} & 55.61 & \textbf{\textcolor{blue}{81.21}} & 78.23 & 70.26 & 81.61 & 54.86 & 43.27 & 65.52 & 63.87 \\
  Gilding Vertex \cite{glid_vertex} & IMP & ResNet-50 & 65.35 & 28.87 & 74.96 & 81.33& 33.88 & 74.31 & 19.58 & 70.72 & 64.70 & 72.30 & 78.68 & 37.22 & 49.64& 80.22 & 69.26 & 61.13 & 81.49 & 44.76 & 47.71 & 65.04 & 60.06 \\
  AOPG \cite{aod_2022_tgrs_dior_r_aopg} & IMP& ResNet-50 & 62.39 & 37.79 & 71.62& 87.63 & 40.90& 72.47& 31.08 & 65.42 & 77.99 & 73.20 & 81.94 & 42.32& 54.45 & 81.17 & 72.69 & 71.31& 81.49 & 60.04 & \bfseries 52.38 & \bfseries 69.99 & 64.41 \\
  DODet \cite{aod_2022_tgrs_dodet} &IMP & ResNet-50 & 63.40 & 43.35 & 72.11 & 81.32 & 43.12 & 72.59 & 33.32 & \bfseries 78.77 & 70.84 & 74.15 & 75.47 & \bfseries 48.00 & 59.31 & \bfseries 85.41 & 74.04 & \bfseries 71.56 & 81.52& 55.47 & \textbf{\textcolor{red}{51.86}} & \textbf{\textcolor{blue}{66.40}} & 65.10 \\
  \hline
  Oriented R-CNN & MAE & ViT-B & 81.04 & 41.86 & 80.79 & 81.39 & 44.83 & 78.35 & 35.12  & 67.67 & 84.85 & 75.44 & 80.80 & 37.65 & 59.33 & 81.15 & 78.70 & 62.87 & 89.83 & 56.17 & 49.87 & 65.36 & 66.65 \\
  Oriented R-CNN & MAE & ViTAE-B & \bfseries 81.30 & 46.73 & 81.09 & 87.62 & 47.38 & 79.79 & 31.99  & 69.72 & 86.71  & 76.23 & 82.13 & 42.47 & 60.45 & 81.20 & 80.11 & 62.75 & 89.75 & 64.56 & 50.77 & 65.33 & 68.40 \\
 Oriented R-CNN & MAE & ViT-B + VSA & \textbf{\textcolor{blue}{81.20}} & 50.68 & 80.95 & 87.41 & 51.27 & \bfseries 80.87 & 34.61  & \textbf{\textcolor{blue}{76.40}} &88.32  & 78.21 & 83.31 & 45.84 & 64.02 & \textbf{\textcolor{red}{81.23}} & 82.87 & 71.31 & 89.86 & 64.66 & \textbf{\textcolor{blue}{50.84}} & 65.81 & 70.48 \\
  Oriented R-CNN & MAE & ViTAE-B + VSA & \textbf{\textcolor{red}{81.26}} & \textbf{\textcolor{red}{52.26}} & \textbf{\textcolor{blue}{81.10}} & \textbf{\textcolor{red}{88.47}} & 51.35 & \textbf{\textcolor{blue}{80.18}} & 37.40  & 75.29 &  \textbf{\textcolor{red}{88.92}} & 77.52 & \textbf{\textcolor{red}{84.33}} & \textbf{\textcolor{blue}{47.31}} & 63.73 & 81.18 & \textbf{\textcolor{blue}{83.03}} & 71.13 & \bfseries 90.04 & 65.01 & 50.81 & 65.82 & 70.81 \\
  \hline
  Oriented R-CNN & MAE & ViT-B + RVSA (Ours) & 81.10 & 48.86 & \bfseries 81.13& \textbf{\textcolor{blue}{88.27}} & 51.02 & 79.91 & \textbf{\textcolor{red}{39.12}} & 74.69 & 88.60 & 77.83 & 83.53 & 46.48 & 64.14 & 81.19 & \bfseries 84.05 & \textbf{\textcolor{red}{71.47}} & 89.97 & \textbf{\textcolor{blue}{65.73}} & 50.58 & 65.62 & 70.67 \\ 
  Oriented R-CNN & MAE & ViT-B + RVSA$^\Diamond$ (Ours) & 80.92 & 49.88 & 81.05 & \bfseries 88.52 & \textbf{\textcolor{red}{51.52}} & 80.17 & \textbf{\textcolor{blue}{37.87}} & 75.96& \textbf{\textcolor{blue}{88.83}} & \textbf{\textcolor{red}{78.46}} & \textbf{\textcolor{blue}{84.01}} & 46.53 & \textbf{\textcolor{blue}{64.18}} & \textbf{\textcolor{blue}{81.21}} & \textbf{\textcolor{red}{84.04}} & \textbf{\textcolor{blue}{71.34}} & \textbf{\textcolor{blue}{89.99}} & 65.41& 50.53 & \textbf{\textcolor{red}{66.49}} & \textbf{\textcolor{blue}{70.85}} \\
  Oriented R-CNN & MAE & ViTAE-B + RVSA (Ours) & 81.05 & \textbf{\textcolor{blue}{50.98}} & 81.04 & 87.90& \textbf{\textcolor{blue}{51.48}} & \textbf{\textcolor{red}{80.63}} & \bfseries 40.96 & 75.91 & 88.73 & \bfseries 78.84 & 83.92 & 46.58 & \textbf{\textcolor{red}{64.57}} & 81.17 & 82.62 & 70.98 & 89.86 & \textbf{\textcolor{red}{65.76}} & 50.36 & 65.72 & \textbf{\textcolor{red}{70.95}} \\ 
  Oriented R-CNN & MAE & ViTAE-B + RVSA$^\Diamond$ (Ours) &  \textbf{\textcolor{blue}{81.20}} & \bfseries 54.71 & \textbf{\textcolor{red}{81.12}} & 88.13 & \bfseries 51.83 & 79.93 & 36.79 & 76.06 & \bfseries 89.23 & \textbf{\textcolor{blue}{78.30}} & \bfseries 84.46 & 47.29 & \bfseries 65.01 & 81.19 & 82.17 & 70.69 &  \textbf{\textcolor{red}{90.03}} & \bfseries 66.75 & 50.73 & 65.40 & \bfseries 71.05 \\
  \hline
\end{tabular}
  }
  \begin{tablenotes}
    \scriptsize
    \item[1] APL: airplane. APO: airport. BF: baseballfield. BC: basketballcourt. BR: bridge. CH: chimney. DAM: dam. ETS: expressway-toll-station. ESA: expressway-service-area. \\ GF: golffield. GTF: groundtrackfield. HA: harbor. OP: overpass. SH: ship. STA: stadium. STO: storagetank. TC: tenniscourt. TS: trainstation. VE: vehicle. WM: windmill.
  \end{tablenotes}
  \end{threeparttable}
  \label{det_dior}
\end{table*}

\subsubsection{Comparison Against State-of-the-art Methods}
We compare the proposed methods with some recently proposed and so far the most advanced methods, and the results are presented in Table \ref{det_dota}-\ref{det_dior}. The top three scores in each metric are marked by \textbf{bold}, \textbf{\textcolor{red}{red}} and \textbf{\textcolor{blue}{blue}}, respectively. On the DOTA-V1.0 dataset, we list the results of single-scale training and multi-scale training, separately. As can be seen, when training on a single scale, our models have advantages in most categories. Concretely, the proposed models perform the best in five classes and surpass the previous best method by about 1\% mAP. In addition, we can also observe that our RVSA models greatly outperform corresponding baselines: ViT-B and ViTAE-B, and achieve higher accuracies compared with the VSA models because of the additional rotation mechanism. In the more competitive multi-scale setting, our models still win the first place in a total of four categories. Owing to the greatly improved detection results in some challenging categories such as the roundabout and harbor, our model sets a new SOTA on DOTA-V1.0, i.e., an mAP of 81.24\%, outperforming all previous methods. 

On the more challenging DIOR-R dataset, the proposed models perform the best in eleven categories. Specifically, we find that the proposed models achieve surprisingly good results in some categories such as the airport, where the performance is even improved over 10\% mAP compared to existing methods. As a result, our models achieve the best performance and significantly outperform the second place by 5\% mAP. When comparing with the original full attention and the VSA variant, our RVSA further improves model performances, which are similar to the conclusion in the experiments of DOTA-V1.0. In addition, we can see that RVSA performs better than RVSA$^\Diamond$ on the DOTA-V1.0 dataset, while RVSA$^\Diamond$ seems to prefer the DIOR-R dataset. These results are consistent with previous observations (See Table \ref{complexity}). It is noteworthy that we successfully demonstrate the possibility of establishing strong plain ViT baselines: ViT-B + VSA and ViTAE-B + VSA, which set new SOTA on DOTA-V1.0 \cite{dota1} and DIOR-R \cite{aod_2022_tgrs_dior_r_aopg} datasets and outperform all previous models, including the model that is even trained with the 1 million labeled RS data, e.g., RSP-ViTAEv2-S \cite{wang_rsp_2022}. In addition, after integrating with the rotation mechanism, their performances can be improved further. We also provide some detection results in Figure \ref{det_vis}, where the first two rows are the results of ViTAE-B + RVSA with multiscale training and testing on DOTA-V1.0, and the remained rows are the results of ViTAE-B + RVSA$^\Diamond$ with multiscale training and testing on DIOR-R. As can be seen, based on the ViTAE backbone equipped with the proposed RVSA modules, Oriented R-CNN can accurately recognize and locate various artificial or natural objects with high confidence no matter in city and rural scenes or dense and sparse distributions. The above qualitative and quantitative results demonstrate the superiority of the proposed RVSA and its effectiveness in advancing plain ViTs toward RS foundation models.

\subsection{Remote Sensing Scene Classification}
\subsubsection{Dataset}
We also evaluate the proposed models for the scene classification task on the UC Merced Land Use (UCM) dataset \cite{ucm}, the Aerial Image Dataset (AID) \cite{aid}, and the famous scene classification benchmark constructed by Northwestern Polytechnical University, called the NWPU dataset \cite{asr_review}. Their details are shown in Table \ref{cls_datasets}.

\begin{table}[t]
  \scriptsize
  \caption{The details of different scene classification datasets.}
  \newcommand{\tabincell}[2]{\begin{tabular}{@{}#1@{}}#2\end{tabular}}
  \centering
  \begin{tabular}{l|cccc}
  \hline
  Dataset  & Number of Sample & Number of Category & Image Size \\
  \hline
  UCM & 2,100 & 21 & 256 $\times$ 256 \\
  AID  & 10,000 & 30 & 600 $\times$ 600\\
  NWPU  & 31,500 & 45 & 256 $\times$ 256 \\
  \hline
\end{tabular}
  \label{cls_datasets}
\end{table}

\subsubsection{Implementation Details and Experimental Settings}
Following \cite{mblanet_2021_asr}, we consider five settings, including UCM-55, AID-28, AID-55, NWPU-19, and NWPU-28 to comprehensively evaluate the proposed models. Here, the suffix $-mn$ means $10\times m\%$ samples are used for training, while the others are for testing. We change the neuron number of the last linear layer to match the number of categories in each dataset. Note that we use the same pretrained models as those in the experiments for object detection and follow the hyper-parameter settings in Section~\ref{subsubsec:implementationdetails} during finetuning on these datasets. We use the top-1 accuracy as the evaluation metric. All experiments are repeatedly conducted five times and we record the average values.

\subsubsection{Experimental Results}

\begin{table*}[t]
  \caption{Results of different methods for scene classification.}
  \newcommand{\tabincell}[2]{\begin{tabular}{@{}#1@{}}#2\end{tabular}}
  \centering
  \resizebox{0.6\linewidth}{!}{
  \begin{tabular}{l|l|l|ccccc}
  \hline
  Pretrain & Backbone & Method & UCM-55 & AID-28 & AID-55 & NWPU-19 & NWPU-28 \\
  \hline
  IMP & VGG-16 & LSENet \cite{lse_2021_asr} & 98.53 & 94.41 & 96.36 & 92.23 & 93.34 \\ 
  IMP & ResNet-50 & $\text{F}^2$BRBM \cite{asr_2021_jstars_f2brbm} & 98.64 & 96.05 & 96.97 & 92.74 & 94.87\\
  IMP & ResNet-50 & GRMANet \cite{asr_2022_tgrs_grmanet} &99.29 & 95.43 & 97.39 & 93.19 & 94.72 \\
  IMP & ResNet-101 & EAM \cite{asr_2021_grsl_eam} & 98.81 & 94.26 & 97.06 & 91.91 & 94.29 \\
  IMP & ResNet-101 & MSANet \cite{asr_2021_jstars_msanet}   & 97.80 & 93.53 & 96.01 & 90.38 & 93.52 \\
  ASP \cite{long_2022_asp} & ResNet-101 & --- & --- & 95.40 & --- & --- & 94.20  \\
  IMP & DenseNet-121 & MGML-FENet \cite{mgmlnet_asr_2021_featurepartioning}& --- & 96.45 & \bfseries 98.60 & 92.91& 95.39 \\
  IMP & MobileNet-V2 \cite{mobilenetv2} & RBFF \cite{asr_2021_icip_rbff} & 95.83 & 91.02 & 93.64 & 84.59 & 88.05 \\
  IMP & ViT-B & --- & 99.15 &  93.81  &  96.08  &  90.96  & 93.96 \\
  IMP & Swin-T & --- &  99.43 &  96.55  &98.10  & 92.73   & 94.70 \\
  CSPT \cite{zhang_2022_cspt}& ViT-B & ---  & --- &  96.75 & --- & --- & 95.11 \\
  CSPT & ViT-L &  ---  & --- &  96.30 & --- & --- & 95.62 \\
  RingMo \cite{ringmo} & ViT-B & --- & --- & 96.54 & 98.38 & 93.46 & 95.35 \\
  RingMo \cite{ringmo} & Swin-B & --- & --- & 96.90 & 98.34 & 94.25 & 95.67 \\
  IMP & ViTAEv2-S & ---  & 99.43 & 96.61 & 98.08 & 93.90 & 95.29\\
  RSP & ViTAEv2-S & ---  & \textbf{\textcolor{blue}{99.62}} & 96.91 & 98.22 & \textbf{\textcolor{blue}{94.41}} & 95.60\\
  \hline
MAE & ViT-B & --- & \bfseries 99.81 & \bfseries 97.47 & \textbf{\textcolor{red}{98.56}} & \bfseries 94.56 & \textbf{\textcolor{red}{95.78}} \\
MAE & ViTAE-B & --- & 99.60 & \textbf{\textcolor{red}{97.20}} & 98.42 & \textbf{\textcolor{red}{94.43}} & \bfseries 95.82 \\
MAE & ViT-B + VSA & --- & 99.41 & 96.85 & 98.30 & 93.74 &  95.29 \\
MAE & ViTAE-B + VSA & --- & 99.43 & 96.90  & 98.34 & 93.98 & 95.53 \\
  \hline
  MAE & ViT-B + RVSA & --- &  \textbf{\textcolor{red}{99.70}} & 96.92 & 98.33 & 93.79 & 95.49 \\
  MAE & ViT-B + RVSA$^\Diamond$ & --- & 99.58 & 96.86 & 98.44 & 93.74 & 95.45 \\
  MAE & ViTAE-B + RVSA & ---& 99.56 & \textbf{\textcolor{blue}{97.03}} & 98.48 & 93.93 & \textbf{\textcolor{blue}{95.69}} \\
  MAE & ViTAE-B + RVSA$^\Diamond$& --- & 99.50 &  97.01 & \textbf{\textcolor{blue}{98.50}} & 93.92 & 95.66 \\
   \hline
\end{tabular}
  }
  \label{cls_acc}
\end{table*}

Table \ref{cls_acc} summarizes the scene classification results of different models in the above five settings. At this task, the MAE pretrained ViT-B obtains the best on most settings since all tokens are attended in MHSA for scene recognition. It can be seen that our RVSA models are superior to previous methods in three settings including UCM-55, AID-28, and NWPU-28. These comparison methods include specially designed scene classification methods that use different backbone networks such as ResNet, VGG, DenseNet, and MobileNet, as well as the recently proposed pretraining-based methods with advanced vision transformers. While in other settings, our models still perform to be competitive with RSP-ViTAEv2-S, which is also pretrained on MillionAID. Compared to VSA methods, we notice that our methods mainly perform worse in the NWPU-19 setting. It is because the proposed RVSA needs a certain amount of training data to learn the optimal window configurations, while NWPU-19 has relatively small-scale training data. When expanding the training set as in NWPU-28, besides ViT-B + VSA and ViTAE-B + VSA, our models can even surpass RSP-ViTAEv2-S and achieve comparable performance.

\subsection{Remote Sensing Semantic Segmentation}
We also transfer the pretrained models to the semantic segmentation task. Besides the Potsdam\footnote{https://www.isprs.org/education/benchmarks/UrbanSemLab/2d-sem-label-potsdam.aspx} and iSAID \cite{isaid} datasets following \cite{wang_rsp_2022}, we also adopt a recently proposed domain adaptation segmentation dataset --- LoveDA \cite{loveda}. Here, we do not distinguish between urban and rural areas and directly utilize the official division for general segmentation.

\subsubsection{Dataset}

\begin{table}[t]
  \caption{The details of different semantic segmentation datasets.}
  \newcommand{\tabincell}[2]{\begin{tabular}{@{}#1@{}}#2\end{tabular}}
  \centering
  \resizebox{\linewidth}{!}{
  \begin{tabular}{l|ccccc}
  \hline
  Dataset  & Training & Validation & Testing & Category & Image Size \\
  \hline
  Potsdam & 24 & --- & 14 & 6 & 6,000 $\times$ 6,000 \\
  iSAID  & 1411 & 458 & 937 & 15 & 800 $\times$ 800 $\sim$ 4,000 $\times$ 13,000 \\
  LoveDA  & 2522 & 1669 & 1796 & 7 & 1,024 $\times$ 1,024 \\ 
  \hline
\end{tabular}
  }
  \label{seg_datasets}
\end{table}

The categories in iSAID are completely the same as DIOR-R since the two datasets share the same set of scenes while targeting different tasks. The details of these datasets for semantic segmentation are shown in Table \ref{seg_datasets}.

\subsubsection{Implementation Details and Experimental Settings}
Most of the experimental settings are the same with \cite{wang_rsp_2022}. Considering that there are more parameters to be trained in larger models, we increase the training iterations to 160k. Besides, the layer decay rate is 0.9. Following \cite{swint,xu2021vitae,wang_rsp_2022}, the UperNet \cite{upernet} is employed as the segmentation framework. We separately upsample or downsample the outputs from block 4, block 6, block 8, and block12 to form the required feature pyramid as in BeiT \cite{beit}. For training on LoveDA, patches of 512 $\times$ 512 resolution are randomly cropped from the images as input. We combine the training and validation sets of LoveDA to form a trainval set for training, while the testing set is unchanged. Besides the overall accuracy (OA) computed for the Potsdam dataset following the common protocol in the RS segmentation community, we report the mean of the intersection over union (IoU) for all categories in iSAID and LoveDA. We follow the single scale setting to calculate all metrics for a fair comparison. These experiments are implemented with 2$\sim$4 NVIDIA A100 GPUs. 

\subsubsection{Experimental Results}

\begin{table}[t]
  \scriptsize
  \caption{Results of different methods for semantic segmentation. }
  \newcommand{\tabincell}[2]{\begin{tabular}{@{}#1@{}}#2\end{tabular}}
  \centering
  \resizebox{\linewidth}{!}{
  \begin{tabular}{l|l|l|ccc}
  \hline
  Method & Pretrain & Backbone & Potsdam & iSAID & LoveDA \\ 
  \hline
  FCN \cite{fcn}& IMP & VGG-16 & 85.59 & 41.70 & 46.69 \\
  DANet \cite{danet}& IMP & ResNet-50 & 89.72 & 57.50 & --- \\
  PSPNet \cite{pspnet}& IMP& ResNet-50 & 89.45 & 60.30 & 48.31 \\
  DeeplabV3+ \cite{deeplabv3_p}& IMP & ResNet-50 & 89.74 & 60.80 & 48.31 \\
  Semantic FPN \cite{semantic_fpn} & IMP & ResNet-50 & --- &59.30 & 48.15 \\
 FarSeg \cite{farseg} & IMP & ResNet-50 & --- & 63.70 & 48.15 \\
  FactSeg \cite{ass_2022_tgrs_factseg} & IMP & ResNet-50 & --- & \textbf{\textcolor{red}{64.80}} & 48.94 \\
  UNetFormer \cite{ass_2022_isprs_unetformer} & IMP & ResNet-18 & \textbf{\textcolor{red}{91.30}} & --- & \textbf{\textcolor{blue}{52.40}} \\
  UperNet \cite{upernet} & IMP & ResNet-50 & 90.64 & 61.90 & 51.27 \\
  UperNet & IMP & Swin-T & 91.17 & \textbf{\textcolor{blue}{64.60}} & 50.00 \\
  UperNet & RingMo \cite{ringmo} & Swin-B & \bfseries 91.74 & \bfseries 67.20 & --- \\
  UperNet & RSP & ViTAEv2-S & 91.21 & 64.30 & \textbf{53.02} \\
  \hline
  UperNet &MAE  & ViT-B &  90.32 & 61.40 & 51.03 \\
  UperNet &MAE  & ViTAE-B &  91.05 & 61.77 & 52.58 \\
  UperNet &MAE  & ViT-B + VSA & 90.54 & 63.55 & 51.29 \\
  UperNet &MAE  & ViTAE-B + VSA & 91.12 & 63.92 & 51.82  \\
  \hline
  UperNet &MAE & ViT-B + RVSA & 90.60 & 63.76 & 51.95 \\
  UperNet &MAE & ViT-B + RVSA$^\Diamond$ & 90.77 & 63.85 & 51.95  \\
  UperNet &MAE & ViTAE-B + RVSA & \textbf{\textcolor{blue}{91.22}} & 63.48 & 52.26 \\
  UperNet &MAE & ViTAE-B + RVSA$^\Diamond$& 91.15 & 64.49 & \textbf{\textcolor{red}{52.44}} \\
  \hline
\end{tabular}
  }
  \label{seg_acc}
\end{table}

Table \ref{seg_acc} shows the results of different methods. It can be seen that our models obtain comparable performance to SOTA methods. We acknowledge that the performance of the proposed models on the segmentation task is not as impressive as those on the object detection and scene classification tasks. We attribute it to two reasons. First, we only use the classical segmentation framework UperNet, which can not effectively propagate high-level semantics to the high-resolution features. Therefore, it is unable to carry out an accurate pixel-level understanding as the latest proposed frameworks such as UNetFormer \cite{ass_2022_isprs_unetformer} and FactSeg \cite{ass_2022_tgrs_factseg}. Another reason may be that the vision transformer backbones we adopted have a plain structure whose tokens are directly embedded from 16 $\times$ 16 patches and the feature map resolution keeps 1/16 of the input size.
Compared with the hierarchical structure like Swin and ViTAEv2, which could generate high-resolution feature maps at early stages, the plain ViTs may lose details, which may harm the pixel-level semantic segmentation task. Nevertheless, the proposed RVSA can still advance the plain ViTs and achieve similar performance to RSP-ViTAEv2-S, outperforming basic ViT-B, ViTAE-B and VSA related models, demonstrating its strong ability to capture useful context from the learned rotated varied-size windows.

\subsection{Comparison with Plain ViT on All Tasks}

\begin{table}[t]
  \scriptsize
  \caption{Average results on all datasets in each task.}
  \newcommand{\tabincell}[2]{\begin{tabular}{@{}#1@{}}#2\end{tabular}}
  \centering
  \begin{tabular}{l|ccc}
  \hline
  Model  & Detection & Classification & Segmentation  \\
  \hline
  ViT-B & 71.85  & 97.24 &  67.58 \\
  ViT-B + RVSA  & 74.71 & 96.85 & 68.77 \\
  ViT-B + RVSA$^\Diamond$  & 74.73 & 96.81  & 68.86 \\ 
  \hline
\end{tabular}
  \label{all_task_datasets}
\end{table}

In this section, we compare the proposed ViT models equipped with RVSA/RVSA$^\Diamond$ and plain ViT for all tasks. The plain ViT performs well in the scene classification task but has poor performance in detection and segmentation tasks. This may be because all tokens probably contribute to the description of the scene category. Thus, taking all tokens into consideration for attention calculation is beneficial for learning image-level feature representation for scene classification. In contrast, object detection and semantic segmentation prefer object-level and pixel-level feature representation, where the semantics highly relate to the nearby tokens while the far-away tokens may contribute less. Thus RVSA and RVSA$^\Diamond$ could better balance the contribution of nearby and far away tokens and perform much better than the plain ViT on the two tasks. Here, taking ViT-B as an example, we present the average results of different models on all the datasets of different tasks in Table \ref{all_task_datasets}. The scores are computed by averaging the metrics across datasets in each task, e.g., the average detection score of ViT-B + RVSA is $(78.75+70.67)/2 \approx 74.71$. As can be seen, ViT-B with the proposed RVSA modules performs much better than the baseline on detection and segmentation tasks since they promote the extraction of rich contexts from diverse windows. Besides, RVSA and RVSA$^\Diamond$ perform comparably on average, although they prefer different datasets at different difficulty levels as shown in Table \ref{cls_acc} and \ref{seg_acc}.

\subsection{Finetuning on Fewer Samples}
\begin{figure}[t]
  \centering
  \includegraphics[width=0.7\linewidth]{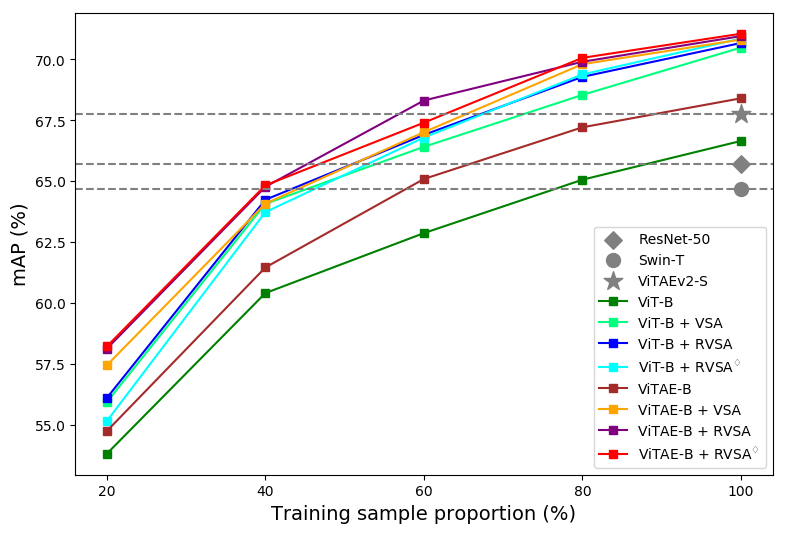}\\
  \caption{Results of different models trained with different amounts of training samples on the DIOR-R dataset. The Oriented R-CNN detection framework is adopted for all models.
  }
  \label{few_shot}
\end{figure}

Data efficiency in transferring is an important ability of foundation models \cite{bommasani2021opportunities, vitae_v2,GPT}. Here, we conduct experiments on the DIOR-R dataset using different amounts of training data. Concretely, we obtain a series of smaller training sets by randomly selecting 20\%, 40\%, 60\%, or 80\% images of the original training set, respectively. Then, we separately finetune the pretrained models on these datasets and evaluate them on the original test set. For comparison, we include some small-scale models such as RSP-ResNet-50, RSP-Swin-T, and RSP-ViTAEv2-S, which are trained with all training samples. Their pretrained weights are obtained from the codebase\footnote{https://github.com/ViTAE-Transformer/ViTAE-Transformer-Remote-Sensing}. Figure \ref{few_shot} shows the results. As can be seen, the proposed models outperform the corresponding ViT-B and ViTAE-B baseline models regardless of the number of training samples. Since we consider the oriented objects of the RSI, the proposed RVSA with the extra learnable rotation mechanism can surpass VSA in most situations. In addition, they achieve comparable performance to Swin-T by using only 40\% training samples, outperform ResNet-50 and Swin-T when 60\% training samples are used, and surpass the strong backbone ViTAEv2-S by using 80\% samples. These findings validate our models' good data efficiency ability in transferring.  Figure \ref{few_shot} also shows that RVSA$^\Diamond$ requires more training samples to perform better than RVSA, as demonstrated in Section~\ref{subsubsec:attentioncomparison}.

\subsection{Visualization of RVSA}

\begin{figure}[t]
  \centering
  \includegraphics[width=\linewidth]{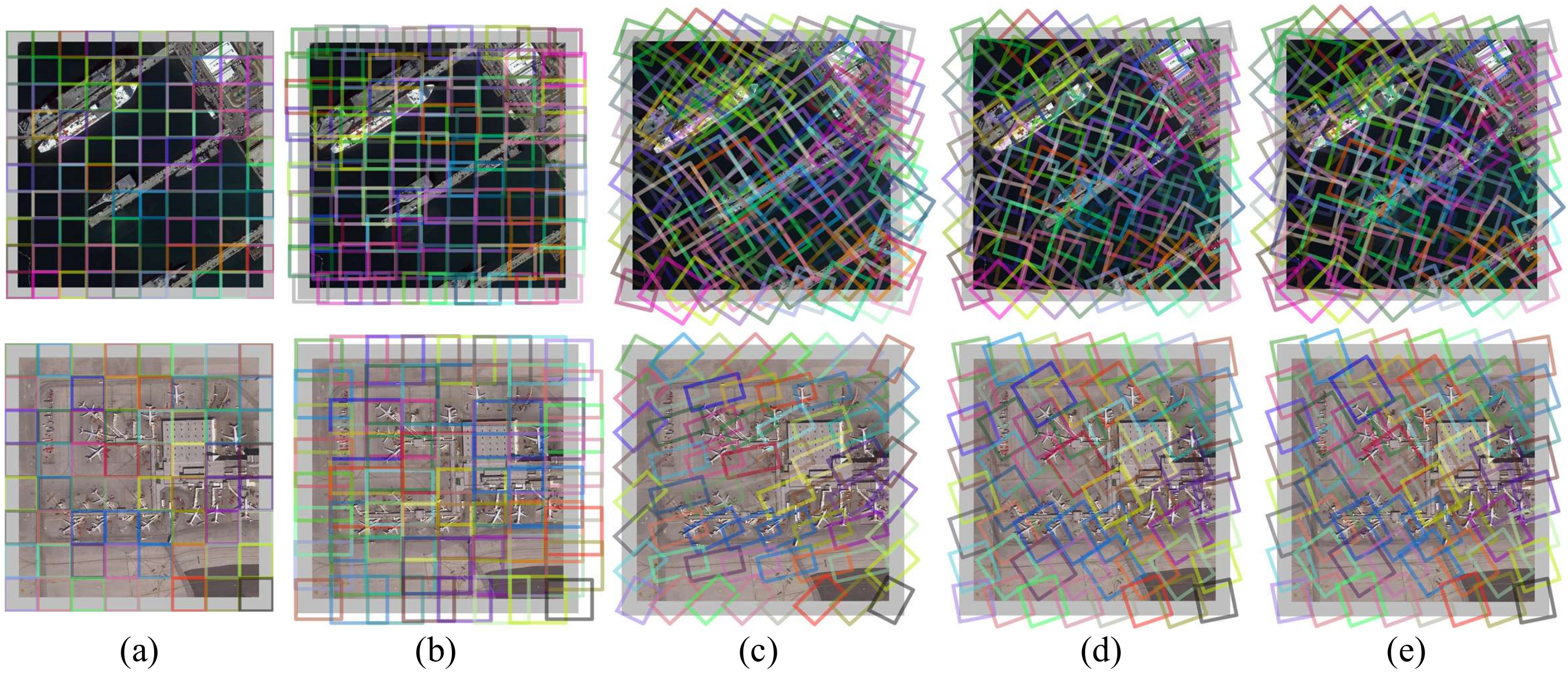}\\
  \caption{Visualization of the generated windows of different attention methods based on ViT-B. (a) Window attention. (b) VSA. (c) RVSA. (d) and (e) are the generated windows of RVSA$^\Diamond$ for the key and value tokens, separately. The images are from the testing sets of DOTA-V1.0 and DIOR-R, respectively.
  }
  \label{window_vis}
\end{figure}

Figure \ref{window_vis} shows the visualization results of generated windows from the attention layer in the penultimate block of different models. These models are based on ViT-B but use different attention methods, including window attention, VSA, RVSA, and RVSA$^\Diamond$. The gray areas denote zero paddings such that the feature size can be divisible by the window size. For example, ViT-B-Win has $(\frac{1024}{16}+6)^2=100$ windows in one head when processing DOTA-V1.0 images (see the first row in Figure \ref{window_vis} (a)) since the zero padding length is 6. As can be seen, the windows generated by VSA can be scaled and shifted to match different objects. Nevertheless, VSA is unable to effectively deal with the arbitrary-oriented objects in RSIs, such as the oriented airplanes in the second row of Figure \ref{window_vis}. By contrast, our RVSA introduces the rotation factor to address this issue, obtaining more diverse windows and promoting the extraction of more rich contexts. It is also noteworthy that the generated windows in one head can well adapt to some airplanes, and each head can produce a different set of windows. Therefore, the airplanes can be covered by the windows in different heads, implying that RVSA can better deal with arbitrary-oriented objects. Compared with RVSA, RVSA$^\Diamond$ further improves the flexibility of generated windows. By comparing (d) and (e) with (c), we can find that there are slight changes in the window shape for key and value tokens, which could be useful when dealing with challenging samples and a large amount of training data available.

\section{Conclusion}
In this paper, we make the first attempt to investigate the potential of the plain ViT towards the RS foundation model. Specifically, we propose a novel rotated varied-size window attention method to advance the performance of plain ViTs. It generates diverse windows at different angles, sizes, shapes, and locations to cover the arbitrary-oriented objects in RSIs and enables to extract rich contexts from the generated windows, thereby promoting the learning of better object representation. We validate the proposed method based on the representative unsupervised pretraining method MAE on typical RS tasks including scene classification, object detection, and semantic segmentation. The results demonstrate the proposed method's superiority and effectiveness in advancing the plain ViT for different tasks. We hope this study could provide valuable insights to the community and inspire future research on developing RS foundation models, especially based on plain ViTs.

\ifCLASSOPTIONcaptionsoff
  \newpage
\fi



\bibliographystyle{IEEEtran}
\normalem
\bibliography{vit_rvsa}

\end{document}